\documentclass[journal]{IEEEtran}

\usepackage{nomencl}
\makenomenclature
\usepackage{cite}
\usepackage{booktabs}
\usepackage{geometry}
\usepackage[ruled,linesnumbered]{algorithm2e}
\usepackage[pdftex]{graphicx}

\usepackage{amssymb}
\usepackage{array}

\usepackage{stfloats}
\fnbelowfloat

\usepackage{url}

\usepackage{color}
\usepackage[colorlinks,linkcolor=blue]{hyperref}

\usepackage[normalem]{ulem} 
\useunder{\uline}{\ul}{}
\usepackage{textcomp} 

\usepackage{comment}
\usepackage[dvipsnames]{xcolor}
\colorlet{zoe}{Orange}
\colorlet{b}{black} 

\usepackage{amsmath}

\newcommand\T{{\hspace{-0pt}\intercal}}

\begin{document}
	%
	\title{Action Planning for Packing Long Linear Elastic Objects into Compact Boxes with Bimanual Robotic Manipulation}
	
	\author{Wanyu Ma, Bin Zhang, Lijun Han, Shengzeng Huo, Hesheng Wang, and David Navarro-Alarcon%
		\thanks{This work is supported in part by the Research Grants Council (RGC) of Hong Kong under grants 14203917.}
		\thanks{W. Ma, B. Zhang, S. Huo and D. Navarro-Alarcon are with The Hong Kong Polytechnic University, Department of Mechanical Engineering, Hung Hom, Kowloon, Hong Kong. (email: wanyu.ma@connect.polyu.hk)}%
		\thanks{L. Han and H. Wang are with the Shanghai Jiatong University, Department of Automation, Shanghai, China.}%
	}

	
	\markboth{IEEE/ASME Transactions on Mechatronics}%
	{Ma \MakeLowercase{\textit{et al.}}: }
	
	\bstctlcite{IEEEexample:BSTcontrol}
	
	\maketitle
	
	\begin{abstract}
		In this paper, we propose a new action planning approach to automatically pack long linear elastic objects into common-size boxes with a bimanual robotic system.
		For that, we developed a hybrid geometric model to handle large-scale occlusions combining an online vision-based method and an offline reference template.
		Then, a reference point generator is introduced to automatically plan the reference poses for the predesigned action primitives. 
		Finally, an action planner integrates these components enabling the execution of high-level behaviors and the accomplishment of packing manipulation tasks.
		To validate the proposed approach, we conducted a detailed experimental study with multiple types and lengths of objects and packing boxes.
	\end{abstract}
	
	
	%
	\IEEEpeerreviewmaketitle
	
	\section{Introduction} \label{section: introduction}
	
	%
	%
	\IEEEPARstart{T}{he} social distancing requirements imposed by the COVID-19 pandemic has forced many businesses to adopt online retail platforms.
	Recent reports \cite{covid_ecommerce} indicate that the pandemic has accelerated the shift away from physical to on-line stores by roughly 5 years. 
	It is predicted that at the current pace [citation], many ageing societies will not have sufficient human workers within a decade to sort and pack products; Human manual packing is simply unsustainable for these communities. 
	A feasible strategy to deal with this record increasing demand for products and diverse commodities (whose heterogeneous properties may vary from compliant, to articulated, to deformable \cite{freichel2020role}) is to use dexterous robots that can automate the soft packing process.
	This approach can also help to optimize the current manual practice in the industry (which tends to use
	excessive packaging that wastes materials), and to improve
	social distancing (as no human workers are needed).
	Our goal in this paper is precisely to develop efficient manipulation strategies that can automate the packing problem of deformable objects.
	
	\begin{figure}[t!]
		\centering
		\includegraphics[width=0.99\columnwidth]{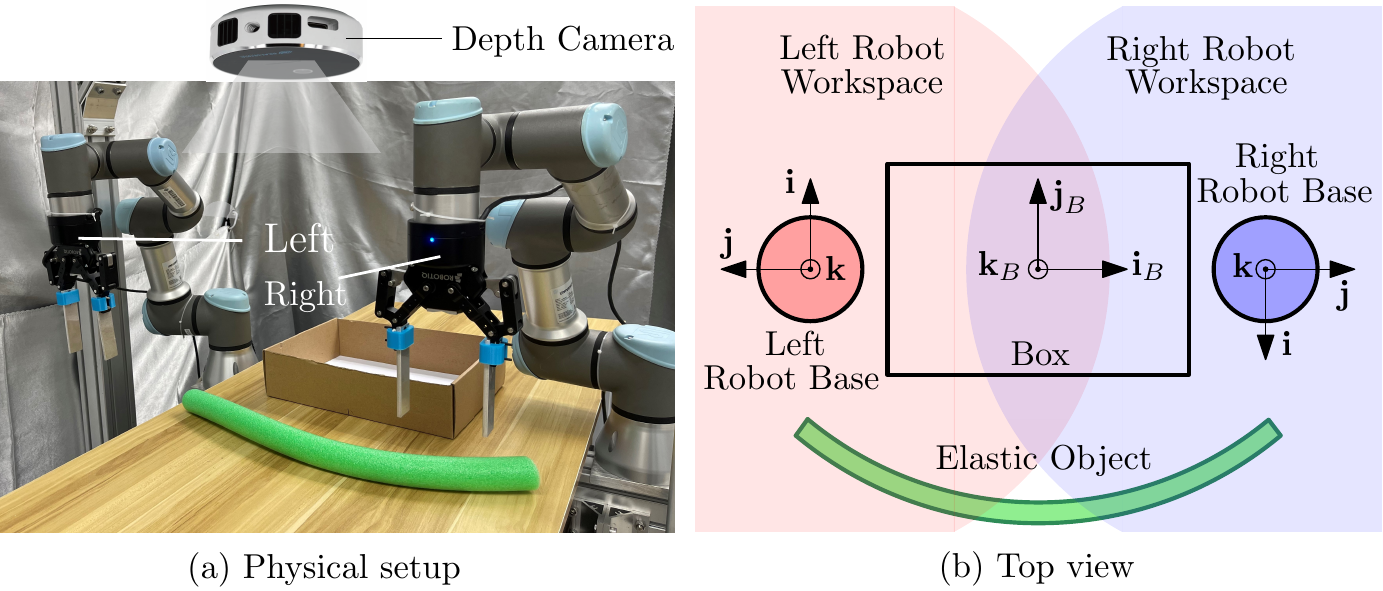}
		\caption{Experimental setup with a box frame $\{F_B\}=\{ \mathbf{i}_B, \mathbf{j}_B, \mathbf{k}_B\}$, two robot arms, the linear elastic object and a top-view camera.}
		\label{experimental setup}
	\end{figure}
	
	To advance in the development of these valuable manipulation skills \cite{zhu_ram,8457261}, in this paper we focus on the challenging problem where a (long) linear elastic object (LEO) needs to be autonomously grasped, shaped/deformed, and placed within a compact box that optimizes its packing space, as depicted in Fig. \ref{experimental setup}.
	There are two main challenges that arise with the automation of this task: (i) Due to the complexity of the shaping task (which is difficult to perform with a single continuous motion), several coordinated actions by collaborative arms are required to effectively deform and place the object within the box; (ii) The typically occluded view from vision sensors during the task leads to partial observations of the manipulated object and the environment (this results in incomplete geometric information that complicates the real-time guidance of the robot's motion).
	
	\subsection{Related Work}
	\subsubsection{Robotic Packing in Logistics}
	Although there has been a strong push towards robotizing the processing
	of product, e.g., with automated guided vehicles in distribution centres \cite{7151854}, packing remains a task entirely performed by human workers. 
	Recently, many methods have been developed for the Amazon Picking Challenge to automatically recognize, collect, and transfer multiple types of products into boxes \cite{schwarz2018fast, yasuda2020packing, yu2016summary}.
	Note that the majority of these methods do not address (and underestimate) large-scale elastic deformations, e.g., those exhibited by linear elastic objects;
	To optimize packing space, shape control is needed to transfer and arrange LEOs into compact boxes.
	The few works that do consider the arrangement of highly deformable materials \cite{DBLP:journals/corr/abs-2012-03385}, do not address shape control and are mostly confined to simple numerical simulations. 
	As the booming e-commerce industry now extends to many non-traditional commodities (e.g., deformable groceries and household products \cite{digital_shopping}), it is essential to develop shape control methods that can deform highly elastic materials and thus save packing space; However, this challenging soft packing problem has not been sufficiently studied in the literature.
	
	\subsubsection{Action Planning for Packing Tasks} In contrast with traditional (low-level) control methods for manipulating soft objects based on \emph{continuous} trajectories \cite{david2014} (i.e., with a single action), the robotic packing of a LEO requires to use \emph{discrete} task planning with multiple (high-level) actions.
	These types of methods decompose and plan the task in terms of a coordinated sequence of action primitives, each of which captures a specific motor behavior (this approach has been used in a wide range of applications, e.g., grasping  \cite{felip2009robust}, soccer \cite{allgeuer2018hierarchical}, assembly \cite{wang2018robot}).
	Action primitives methods have been proposed for packing and object arrangement problems, e.g. \cite{schwarz2017nimbro} develops a controller for robotic picking and stowing tasks based on parametrized motion primitives; \cite{zeng2018learning} proposes a method for manipulating objects into tightly packed configurations by learning pushing/grasping policies; \cite{capitanelli2018manipulation} tackles the problem of reconfiguring articulated objects by using an ordered set of actions executed by a dual-arm robot. 
	Yet, note that the action primitives adopted by these works cannot capture the complex behaviors that are needed to control the shape of a LEO during a packing task.
	
	\subsubsection{Representation of Deformable Objects}
	To visually guide the manipulation task, it is necessary for a controller to have a meaningful representation of the object.
	To this end, researchers have developed a variety of representation methods, e.g., physics-based approaches \cite{kimura2003constructing, essahbi2012soft, 9215039, petit2017using, kaufmann2009flexible} (using mass-spring-damping models and finite element method), based on visual geometric features  \cite{qi2021contour, 9000733, 8676321,laranjeira2020catenary} (using points, angles, curvatures, catenaries, contours moments, etc), or data-driven representations \cite{navarro2018fourier, hu20193, zhu2021vision, 9410363} (using Fourier series, FPFH, PCA, autoencoders, etc).
	Typically, vision-based approaches are strongly affected by occlusions during the task (this is problematic for packing as a top observing camera will have incomplete observations during the task). 
	To deal with this issue, many works have addressed the estimation and tracking of the object's deformation in real-time \cite{tang2018track, jin2022robotic, chi2019occlusion}; Yet, these works only consider with 2D scenarios, hence, are not applicable to our 3D LEO manipulation problem.
	
	
	\subsection{Our Contribution}
	In this paper, we propose: (1) A new hybrid geometric model that combines online 3D vision with an offline reference template to deal with camera occlusions; (2) A reference point planner that provides intermediate targets to guide the high-level packing actions; (3) A cyclic action planner that coordinates multiple action primitives of a dual-arm robot to perform the complex LEO packing task.
	The proposed methodology is original, and its demonstrated capabilities have not (to the best of the authors' knowledge) been previously reported in the literature.
	To validate this new approach, we report a detailed experimental study with a dual-arm robot performing packing tasks with LEOs of various elastic properties.
	
	This paper is organized as follows: Section \ref{section: modeling} presents the mathematical models; Section \ref{section: Hybrid Geometric Model} describes the hybrid geometric model; Section \ref{section: action planner} presents the packing method; Section \ref{section: results} reports the experiments; Section \ref{section: conclusions} gives conclusions.
	
	\section{Modeling}\label{section: modeling}
	Table \ref{nomenclature} presents the key nomenclature used in the paper. 

	\begin{table}[t!]
		\centering
		\caption{Key Nomenclature}
		\begin{tabular}{l m{6.5cm} }
			\toprule[1pt]
			$\hspace{-2mm}$Symbol & Quantity \\
			\specialrule{0.5pt}{1pt}{2pt}
			$\hspace{-2mm}\mathcal{O}(\eta, l_O, d_O)$ & LEO of material $\eta$, length $l_O$ and rod diameter $d_O$.\\
			\specialrule{0.01pt}{1pt}{2pt}
			$\hspace{-2mm}\mathcal{B}(l\hspace{-0.5mm}_B,w\hspace{-0.5mm}_B,h\hspace{-0.5mm}_B)$ & Cuboid container of dimensions $l\hspace{-0.5mm}_B\times w\hspace{-0.5mm}_B\times h\hspace{-0.5mm}_B$.\\
			\specialrule{0.01pt}{1pt}{2pt}
			$\hspace{-2mm}\mathbf{P}^*$ & The offline reference template.
			\\
			\specialrule{0.01pt}{1pt}{2pt}
			$\hspace{-2mm}\mathbf{P}$ & The raw feedback point cloud inside the box.
			\\
			\specialrule{0.01pt}{1pt}{2pt}
			$\hspace{-2mm}\mathbf{P}^{O}$ & The ordered skeleton of the point cloud outside the box. 
			\\
			\specialrule{0.01pt}{1pt}{2pt}
			$\hspace{-2mm}\hat{\mathbf p}_i$ & The corresponding point in $\mathbf{P}^O$ to the $i$th point in $\mathbf{P}^*$.\\
			\specialrule{0.01pt}{1pt}{2pt}
			$\hspace{-2mm}e_{in}$, $e_{out}$ & The shape differences of $\mathbf{P}$ and $\mathbf{P}^O$ to $\mathbf{P}^*$, respectively.\\
			\specialrule{0.01pt}{1pt}{2pt}
			$\hspace{-2mm} e$, ${e}^*$ & Total shape difference and its desired value.\\
			\specialrule{0.01pt}{1pt}{2pt}
			$\hspace{-2mm}\mathbf{x}$ & End-effector feedback pose.\\
			\specialrule{0.01pt}{1pt}{2pt}
			$\hspace{-2mm}\mathbf{x}^*$ & End-effector reference pose. 
			\\
			\specialrule{0.01pt}{1pt}{2pt}
			\hspace{-2mm}$\mathbf{u}$ & 
			End-effector target pose.
			\\
			\specialrule{0.01pt}{1pt}{2pt}
			$\hspace{-2mm}\{ F \}$ & The reference template frame.
			\\
			\specialrule{0.01pt}{1pt}{2pt}
			$\hspace{-2mm}\{ F_O \}$ & The object body frame.
			\\
			\specialrule{0.01pt}{1pt}{2pt}
			$\hspace{-2mm}\Delta h$, $\Delta f$ & Height offsets for hover and fixing action primitives.\\
			\specialrule{0.01pt}{1pt}{2pt}
			$\hspace{-2mm} \delta_l$, $\delta_f$ & Horizontal distances for the reference point generator.\\
			\specialrule{0.01pt}{1pt}{2pt}
			$\hspace{-2mm}\mathbf{p}^{\hspace{-0.3mm}L}\hspace{-0.5mm}$,~ $\hspace{-0.5mm}\mathbf{p}^{\hspace{-0.3mm}G}\hspace{-0.5mm}$,~ $\hspace{-0.5mm}\mathbf{p}^{\hspace{-0.3mm}F}$ & Reference points for object placing, grasping, and fixing.\\
			\bottomrule[1pt]
		\end{tabular}
		\label{nomenclature}
	\end{table}

	\subsection{Geometric Object Modeling}
	In our method, we use an RGB-D camera to capture point clouds of the scene in real-time.
	During the task, the raw point cloud is splitted into two structures: $\mathbf{P}^O$ which represents the object's part to be grasped by the robot, and $\mathbf{P}$ which represents the object's part already packed; Spatially, these two structures correspond to the object's parts outside and inside the box, respectively.
	To provide an intuitive topology that facilitates the LEO's manipulation, the points in $\mathbf P^O$ are ordered along the linear object's centerline. 
	During initialization, the object's length $l_O$ and width $d_O$ are computed from the raw point cloud data.
	
	The offline reference template $\mathbf{P}^*=[\mathbf p_1^*,\ldots,\mathbf p_M^*]\in\mathbb R^{3\times M}$ is a pre-designed geometric curve which represents the final configuration to be given to the object $\mathcal{O}(\eta, l_O,d_O)$ within the box $\mathcal{B}(l_B,w_B,h_B)$.
	Similarly to the raw feedback point cloud, the reference template $\mathbf{P}^*$ is also separated into two parts at the split point $\mathbf{p}_s$, correspongding to $\mathbf{P}$ and $\mathbf{P}^O$. 
	Given the $i$th point $\mathbf p_i^*$ on $\mathbf{P}^*$, we denote its corresponding point at $\mathbf P^O$ as $\hat{\mathbf{p}}_i^*$.
	The length from $\hat{\mathbf{p}}_i^*$ to the end of the object $\mathbf{p}^O_{N}$ equals the length from $\mathbf{p}_i^*$ to $\mathbf{p}_{M}^*$.

\begin{figure}[t!]
		\centerline{\includegraphics[width=\columnwidth]{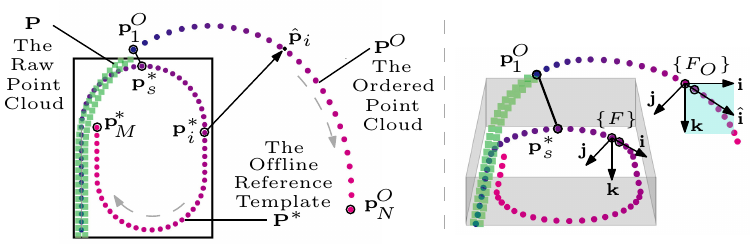}}
		\caption{The the proposed hybrid geometric model of a LEO. Blue indicates the start curve and pink indicates its end. The axes $\mathbf i$, $\mathbf k$ and $\hat{\mathbf i}$ are all co-planar.}
		\label{hybrid-geometric-model}
	\end{figure}

	For the point clouds $\mathbf{P}^*$ and $\mathbf{P}^O$, two important frames, the reference template frame and the object body frame, are defined.
	The reference template frame is denoted as $\{ F \}$ at $\mathbf{p}^*_{i}, i = 1,...,M-1$. 
	Its z-axis is the unit vector vertically pointing to the table. 
	The x-axis is the unit vector poiting to $\mathbf{p}^*_{i+1}$. 
	And the y-axis is determined by the right-hand principle.
	The object body frame is denoted as $\{ F_O \}  =  \{\mathbf{i}, \mathbf{j}, \mathbf{k}\}$ at a point in $[\mathbf{p}_{i}^O, \mathbf{p}_{i+1}^O), i =\hspace{-1mm} 1,...,N - 1$. 
	Similarly, its z-axis is the unit vector vertically pointing to the table. 
	Then, we introduce the unit tangent vector $\hat{\mathbf{i}}$ pointing from $\mathbf{p}_{i}^O$ to $\mathbf{p}_{i+1}^O$.
	The y-axis $\mathbf{j}$ is orthogonal to the plane $\mathbf{k}-\hat{\mathbf{i}}$.
	Note that, since the object is not in a plane parallel to the table, $\hat{\mathbf{i}}$ is not always orthogonal to $\mathbf{k}$. 
    So the real x-axis of  $\{ F_O \}$ is determined by $\mathbf i = \mathbf j  \times  \mathbf k$.
	These definations are depicted in Fig. \ref{hybrid-geometric-model}.
	
	To compute shape difference $e$ between the feedback point cloud and the reference template, we must introduce the inside and outside the box errors between these two structures.
	For that, we define as $e_{in}$ the average minimum Euclidean distance (i.e., the similarity \cite{tian2017geometric}) between $\mathbf P$ and the first $s$ points of $\mathbf P^*$, and define as $e_{out}$ the average Euclidean distance between the other $M-s$ points of $\mathbf P^*$ and their corresponding points $\hat{\mathbf p}_i$.
	The total shape difference $e$ is computed as a weighted combination of these two errors:
	\begin{equation}
		e = w e_{in} + (1-w) e_{out}
	\end{equation}
	for $w = \frac{s}{M}$ as normalization weight.
	This metric quantifies the accuracy of the automatic shaping process. 
	Note that, $e_{in}$ and $e_{out}$ are respectively averaged based on the number of points in $\mathbf P$ and $\mathbf P^O$, thus, they represent the distances between two pairs of points.
	Therefore, these errors are not significantly influenced if some feedback points are lost due to occlusion.
	Besides, $e$ is weighted based on the number of points at the two sides of $\mathbf p^*_s$; This gurantees that the contribution of $e_{in}$ and $e_{out}$ is normalized.
	As the feedback point clouds of the LEO represent points over its surface whereas the reference template represents a centerline, therefore, the error $e$ will ideally converge to a desired value ${e}^*= \frac{d_O}{2}$, i.e., half width of the object.

	\subsection{Action Planning}
	The robotic system considered in this study is composed of two end-effectors, one with an active grasp role and the other with an assistive fix role. 
	It is assumed that the end-effectors are vertically pointing towards the box's plane.
	The configuration of the robotic arms is represented by a four degrees-of-freedom (DOF) pose vector $\mathbf x=[x,y,z,\theta{]}^\T$ (comprised of position and orientation coordinates) and one DOF for the gripper's open/close configuration.
	The initial pose $\mathbf x(t_0)$ of the robot arms is assumed to be above the box and object.
	
	The action planner that coordinates multiple action primitives performed by the robot arms is modeled with a classic state machine \cite{hudson2019learning}. 
	This state machine is represented by the tuple $(S,T,A,G,R)$, whose elements are defined as:
	\begin{itemize}
		\item $A$: collection of action primitives for the end-effectors' 4-DOF pose.
		\item $G$: collection of action primitives for the grippers' 1-DOF open/close configuration.
		\item $R$: collection of the robot active and assitant roles.
		\item $S$: collection of action planner states, each represented by a robot movement. 
		\item $T$: state transition function.
	\end{itemize}
	Aiming at recycling the modules of the state machine, the designed action primitives compose a periodic action planner \textit{loop} that enables the robot to automatically perform the complex packing task.
	
	\subsection{Framework Overview}
	\begin{figure}[t!]
		\centerline{\includegraphics[width=\columnwidth]{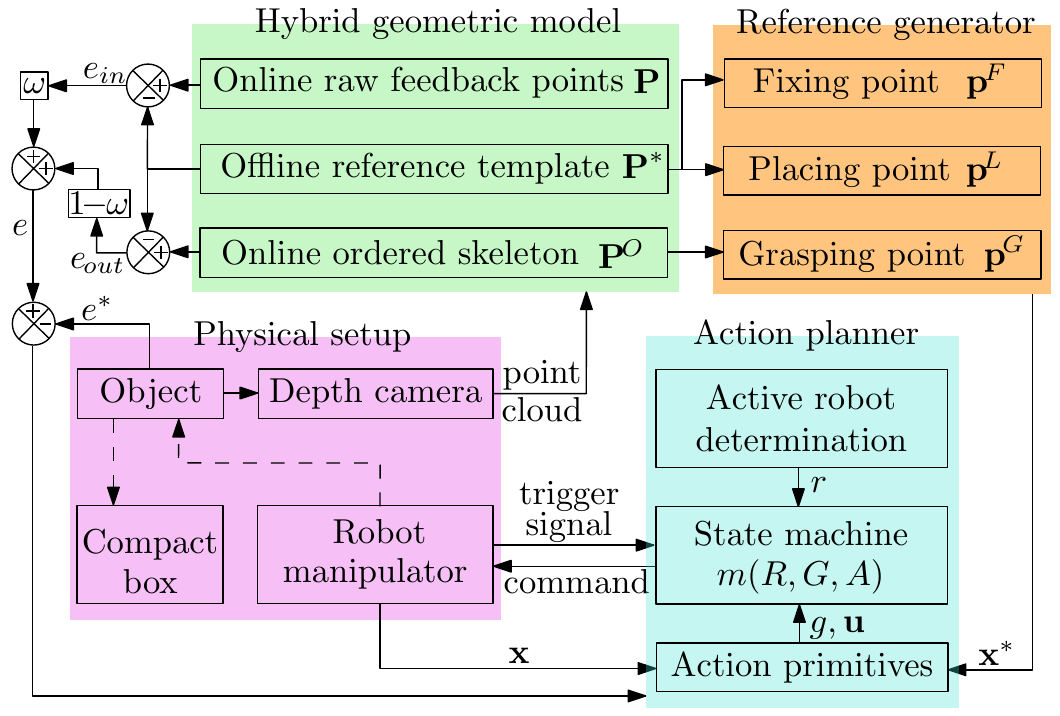}}
		\caption{The framework of the proposed approach for packing long LEOs into common-size boxes. Solid lines indicate data transmission, and dashed lines indicate physical contact.}
		\label{framework}
	\end{figure}
	
	The overview of the proposed automatic packing approach is shown in Fig. \ref{framework}. 
	It is composed of a hybrid geometric model (green block), a reference point generator (orange block), and an action planner (blue block).
	The hybrid geometric model provides a robust representation of the object by combining an online part ($\mathbf P$ and $\mathbf P^O$) and an offline part ($\mathbf P^*$).
	The reference point generator computes reference poses $\mathbf{x}^*$ for the robot to perform grasping, placing, and fixing the object.
	The action planner commands the execution of the task based on a series of action primitives.
	The robotic platform (pink block) receives and executes the kinematic motion command (i.e., the target poses and gripper configurations) from the action planner, and returns an end flag to the control system after its completion.
	The state machine recycles a perdiodic action planner loop by alternating each robot arm between an active and an assistant role until the task is completed.
	
	\section{Hybrid Geometric Model}\label{section: Hybrid Geometric Model}

	The proposed hybrid geometric model consists of $\mathbf P$ (the raw feedback point cloud inside the box), $\mathbf P^O$ (the ordered skeleton of the object's part outside the box), and $\mathbf P^*$ (the generated offline reference template).  
	It extracts the object's geometry in real time and generates the suitable target shape for packing LEOs, which are preconditions of reference point generation and packing progress measurement.
	On one hand, the reference point generator replaces $\mathbf P$ with the corresponding points ($\mathbf p^*_1$ to $\mathbf p^*_s$) of $\mathbf P^*$ and searches for the reference points in $\mathbf P^O$ and $\mathbf P^*$, to deal with the typical occlusions that result from the grippers blocking the top-view camera.
	On the other hand, the shape difference $e$ is computed as the combination of $e_{in}$ (the distance from $\mathbf P$ to the template points $\mathbf p^*_i$, $i = 1, \ldots, s$) and $e_{out}$ (the distance from $\mathbf P^O$ to the template points $\mathbf p^*_i$, $i = s, \ldots, M$), to monitor and quantify the object's packing.

	\subsection{Offline Reference Template}
	\begin{figure}[t!]
		\centering
		\includegraphics[width=\columnwidth]{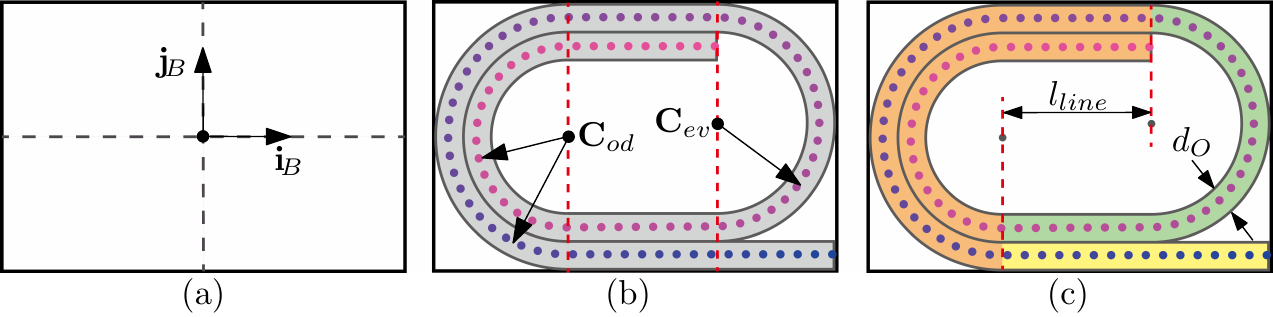}
		\caption{Target shape $Spiral$ of a LEO in a box. (a) shows the box's bottom. The points in gradients from pink to lue resresents $\mathbf{P}$. (b) illustrates how $Spiral$ is constructed with straight segments (between red dash lines) and two sets of concentric semicircles (the centers are $\mathbf{C}_{od}$ and $\mathbf{C}_{ev}$). (c) illustrates the beginning segment (yellow) and periodic parts (orange and green) in $Spiral$. }
		\label{fig:object-model}
		\vspace{-0.2cm}
	\end{figure}
	
	The offline reference template is needed to perform the packing task, as it provides the final target shape of the object and replaces the occluded parts with its offline 3D points. 
	To optimize packing space, the taget shape for the long LEO is designed in the form of a modified spiral, which is composed of straight segments and concentric semicircles, as shown in Fig. \ref{fig:object-model}. 
	This target configuration is separated into periodic and aperiodic parts.
	The former consists of a semicircle followed by a straight segment; The latter only represents the beginning straight segment of the curve.
	Given a box-object pair, the maximum number of action planner loops (which equals to the number of grasps needed to complete the task) is one more than the total number of semicircles, i.e., $\lfloor \frac{w_B}{d_O}\rfloor$, where $\lfloor \cdot \rfloor$ denotes the rounded down nearest integer operator.
	
	We compute the maximum object length that can be placed in the box with the spiral shape as:
	\begin{equation}
		l_O = l_B - \frac{w_B}{2} + \sum_{j=1}^{\lfloor \tfrac{w_B}{d_O}\rfloor} {\left(l_B - w_B+ \frac{d_O}{2} + \pi \frac{w_B-d_Oj}{2} \right)}
		\label{capacity}
	\end{equation}
	Then, we parameterize the centerline of the spiral shape (see Fig. \ref{fig:object-model} (b)) with a normalized length $\lambda = i/M \in[0, 1]$, for $i=1,\dots,M$.
	The parameterized centerline is denoted as $\mathbf P^*(\lambda)$, and the length of its curve is computed as:
	\begin{equation}
		l(\lambda) = \lambda l_O = \sum_{i=1}^{M} \| \mathbf{p}_i^* - \mathbf{p}_{i-1}^* \|_2.
	\end{equation}
	The process for generating the target spiral shape is presented in Algorithm \ref{algorithm3}.
	\begin{algorithm}[t!]
		\small
		\caption{\small The description of the shape $Spiral$}\label{algorithm3}
		\KwIn{the box $\mathcal{B}(l_B, w_B, h_B)$, the object $\mathcal{O} (\eta, l_O, d_O)$}
		\KwOut{the parameterized fomula $\mathbf{P}^*(\lambda)$}
		$\mathbf{C}_{od}(-\frac{l_B}{2}+\frac{w_B}{2}, 0, \frac{d_O}{2})$,
		$\mathbf{C}_{ev}(\frac{l_B}{2}-\frac{w_B}{2}+\frac{d_O}{2}, \frac{d_O}{2}, \frac{d_O}{2})$\; 
		$l_{line} = l_B-w_B+\frac{d_O}{2}$\;
		$l_{count} = l_B-\frac{w_B}{2}$\;
		$j=0$, $\lambda=0$\;
		\While{$\lambda<1$}{
			\eIf{$0 \leq \lambda l_O <l_B-\frac{w_B}{2}$}{
				$\mathbf{P}(\lambda)=(\frac{l_B}{2},\frac{-w_B+d_O}{2},\frac{d_O}{2}) + (\lambda l_O,0,0)$\;
			}
			{
				$j=j+1$\;
				$\lambda=(l_B-\frac{w_B}{2})/l_O$\;
				$r_{sc}=\frac{w_B}{2}-\frac{d_O}{2}j$\;
				$l_{semicircle}=\pi r_{sc}$\;
				\If{$\lambda l_O-l_{count}<l_{semicircle}$}{
					$\phi = \frac{\lambda l_O-l_{count}}{r_{sc}}$\;
					$\mathbf{P}(\lambda)=\mathbf{C}_{od} - r_{sc}(\sin{\phi},\cos{\phi},0)$, $j$ is odd\;
					$\mathbf{P}(\lambda)=\mathbf{C}_{ev} + r_{sc}(\sin{\phi},\cos{\phi},0)$, $j$ is even\;
					$\lambda=\lambda+\frac{1}{M}$\;
				}
				$l_{count} = l_{count} + l_{semicircle}$\;
				\If{$\lambda l_O-l_{count} \leq l_{line}$}{
					$\mathbf{P}(\lambda)=\mathbf{C}_{od} +(\lambda l_O-l_{count}, r_{sc})$, $j$ is odd\;
					$\mathbf{P}(\lambda)=\mathbf{C}_{ev} -(\lambda l_O-l_{count}, r_{sc})$, $j$ is even\;
					$\lambda=\lambda+\frac{1}{M}$\;
				}
				$l_{count} = l_{count} + l_{line}$\;
			}
		}
	\end{algorithm}

	\subsection{Online 3D Vision}\label{section: perception}
	
	\begin{figure}[t!]
		\centerline{\includegraphics[width=\columnwidth]{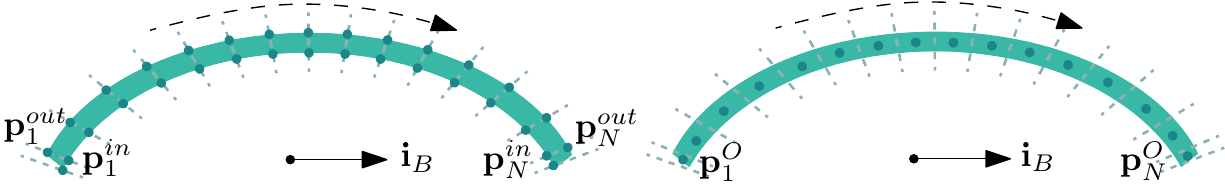}}
		\caption{Point cloud processing: (a) boundary extraction, (b) ordered skeleton.}
		\label{perception}
				\vspace{-0.2cm}
	\end{figure}

	To compute the ordered skeleton $\mathbf P^O$, the point cloud processing algorithm extracts geometric information of the objects in real time.
	Firstly, it smoothens the raw point clouds with a weighted filter \cite{hesterberg1995weighted} and downsamples it to optimize its computational cost.
	Next, it detects the boundaries of the object from the point cloud. 
	For that, we introduce a polar coordinate system with the origin at center of the box and its axis define along $\mathbf i_B$ (as depicted in Fig. \ref{experimental setup}); Then, we segment the object into $N$ sections by rotating (clockwise) a ray starting from $\mathbf i_B$ around the center, and with a fixed angle interval (see Fig. \ref{perception}). 
	Along each ray, we search for nearest and farthest points in the raw point cloud feedback, which we denote as $\mathbf p_i^{in}$ and $\mathbf p_i^{out}$, respectively; These points will serve as the components of the inner boundary and outer boundary of the object. 
	Lastly, the LEO's ordered skeleton $\mathbf P^O$ is constructed by computing the mean of the raw feedback points between two adjacent rays. 
	The length $l_O$ and width $d_O$ of the linear object are intuitively calculated as follows:
	\begin{equation}
		l_O= \hspace{-0.5mm}\sum_{i=2}^{N} \left\|\mathbf{p}_{i}^O - \mathbf{p}_{i-1}^O\right\|_2,\ 
		d_O=\frac{1}{N}\sum_{i=1}^{N} \left\|\mathbf{p}_i^{out} - \mathbf{p}_{i}^{in}\right\|_2.
		\label{equ:geometry}
	\end{equation}
	
	\section{Automatic Packing Method}\label{section: action planner}
	\subsection{Reference Points Generator}\label{section Reference Points Generator}

	
	Our proposed manipulation method recycles a periodic action planner loop, which is composed of various high-level behaviors.
	To execute these behaviors, three types of points are planned, namely, the grasping reference $\mathbf p^G$, the placing reference $\mathbf p^L$ and the fixing reference $\mathbf p^F$.
	The reference point generator (the orange block in Fig. \ref{framework}) constructs the reference pose $\mathbf x^*=[x^*,y^*,z^*,\theta^*]^\T$ for the robot based on the object body frame $\{F_O\}$ (which is computed from $\mathbf P^O$ for $\mathbf p^G$) and the reference template frame $\{F\}$ (which is computed from $\mathbf P^*$ for $\mathbf p^L$ and $\mathbf p^F$). 
	Fig. \ref{hybrid-geometric-model} conceptually depicts these frames.


	\begin{figure}[t!]
		\centering
		\includegraphics[width=\columnwidth]{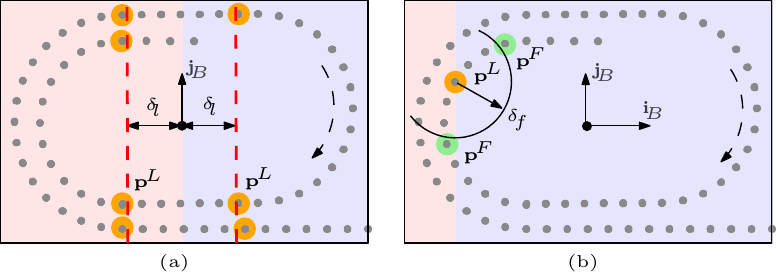}
		\caption{Reference point generator. The offline reference template starts from a corner of the box and the point index clockwise increases. (a) the candidate placing points (orange), (b) the cadidate fixing point (green) given a placing point (orange). Red and blue regions respectively indicate the workspaces of the left and right arms.}
		\label{path-planning}
				\vspace{-0.3cm}
	\end{figure}

	The points $\mathbf{p}^{L}$ represent the positions within the box where the LEO is to be placed by robot.
	These points are a subset of the offline reference template $\mathbf P^*$ and are defined as $\mathbf{p}^{L} = \mathbf{p}_{k}^*$, for $k=1,\ldots,M$ as the index for the points in the template. 
	The index $k$ is chosen such that $\mathbf{p}_{k}^*$ is approximetly at a distance $\delta_L$ from the axis $\mathbf{j}_B$ (i.e., along the red dash lines shown in Fig. \ref{path-planning} (a)).
	Depending on which robot arm plays the active packing role, $\mathbf{p}_{k}^*$ is automatically chosen from the left or right to $\mathbf{j}_B$.
	The points $\mathbf{p}^{G}$ indicate the positions on the object to be grasped by the robot.
	These points are the corresponding points to $\mathbf{p}^L$ on the object's skeleton $\mathbf P^O$, and are computed as $\mathbf{p}^{G}=\hat{\mathbf p}_k$, for $k=1,\ldots,M$.
	The LEO's shaping behavior is achieved by driving $\mathbf p^G$ into $\mathbf p^L$.
	
	In contrast with inelastic linear deformable objects such as ropes or cords \cite{tang2018track}, LEOs have an intrinsitc elastic energy that restores its shape to the original configuration.
	Thus, to steadily place the object in the box requires the assitant robot arm to fix the deformed object at a point $\mathbf p^F$ while the active arm moves to a new gasping point $\mathbf p^G$.
	To compute $\mathbf p^F$, our method first obtains two candidate points along the object that are approximately at a distance $\delta_f$ from the placing point $\mathbf p^L$.
	The fixing point is selected at the same side of the assitant robot arm with respect to the active robot, see Fig. \ref{path-planning} (b).
	
	\subsection{Action Primitives}\label{section:action primitives}
	
	As modelled in Sec. II-B, our method adopts two types of action primitives (for grippers and end-effectors) to compose high-level manipulation behaviors.
	The collection of action primitives for the 1-DOF grippers is as follows: 
	\begin{equation}
		G = \{\textit{Open}, \textit{Close}\} = \{g_1, g_2\}
	\end{equation}
	where the flags $g_1 = 1$ and $g_2 = 0$ define the closing and opening action of grippers, respectively.
	
	The collection of five active primitives for the robotic end-effectors is as follows: 
	\begin{align}
		A &= \{ \textit{Hover}, \textit{Approach}, \textit{Fix}, \textit{Leave}, \textit{Reset}\} \nonumber \\
		&= \{ a_1, a_2, a_3, a_4, a_5\}
	\end{align}
	These end-effector action primitives are defined as follows:
	\begin{itemize}
		\item[$a_1$:] \textit{Hover}. The robot moves and stops above the reference point by an offset $\Delta h$. 
		With this action, the robot is commanded with an end-effector target pose $\mathbf u = \left[x^*, y^*, z^*+\Delta h, \theta^* \right]^{\T}$.
		This action is needed to avoid collisions with the object, and is done as a preparation step to perform fix and grasp actions.
		\item[$a_2$:] \textit{Approach}. The robot  descends to $z^*$ (viz. the height of the object's centerline).
		With this action, the robot is commanded with an end-effector target pose $\mathbf{u} = \left[x , y , z^*, \theta \right]^{\T}$.
		This action, in combination with \textit{Hover}, is needed to grasp and/or place the object by changing the gripper's configuration.
		\item[$a_3$:] \textit{Fix}. The robot descends to the object's surface, whose height is denoted by $z^* + \Delta f$.
		With this action, the robot is commanded as $\mathbf u = \left[x,y,z^*+\Delta f,\theta \right]^{\T}$.
		This motion is needed to push the deformed elastic object and keep it inside the box, 		thus, preventing it from returning to its original shape. 
		\item[$a_4$:] \textit{Leave}. The robot returns to its initial height.
		With this action, the robot is commanded with an end-effector target pose $\mathbf u = \left[x , y , z(t_0) , \theta \right]^{\T}$.
		This action is needed to provide the robot with an obstacle-free region above the box's packing workspace.
		\item[$a_5$:] \textit{Reset}. The robot returns to its initial pose $\mathbf u = \mathbf x(t_0) $.
		This action is needed to visually observe the object with the top-view camera.
	\end{itemize}

	
	\subsection{State Machine}\label{section: state machine}
	\begin{figure}[t!]
		\centering
		\includegraphics[width=0.9\columnwidth]{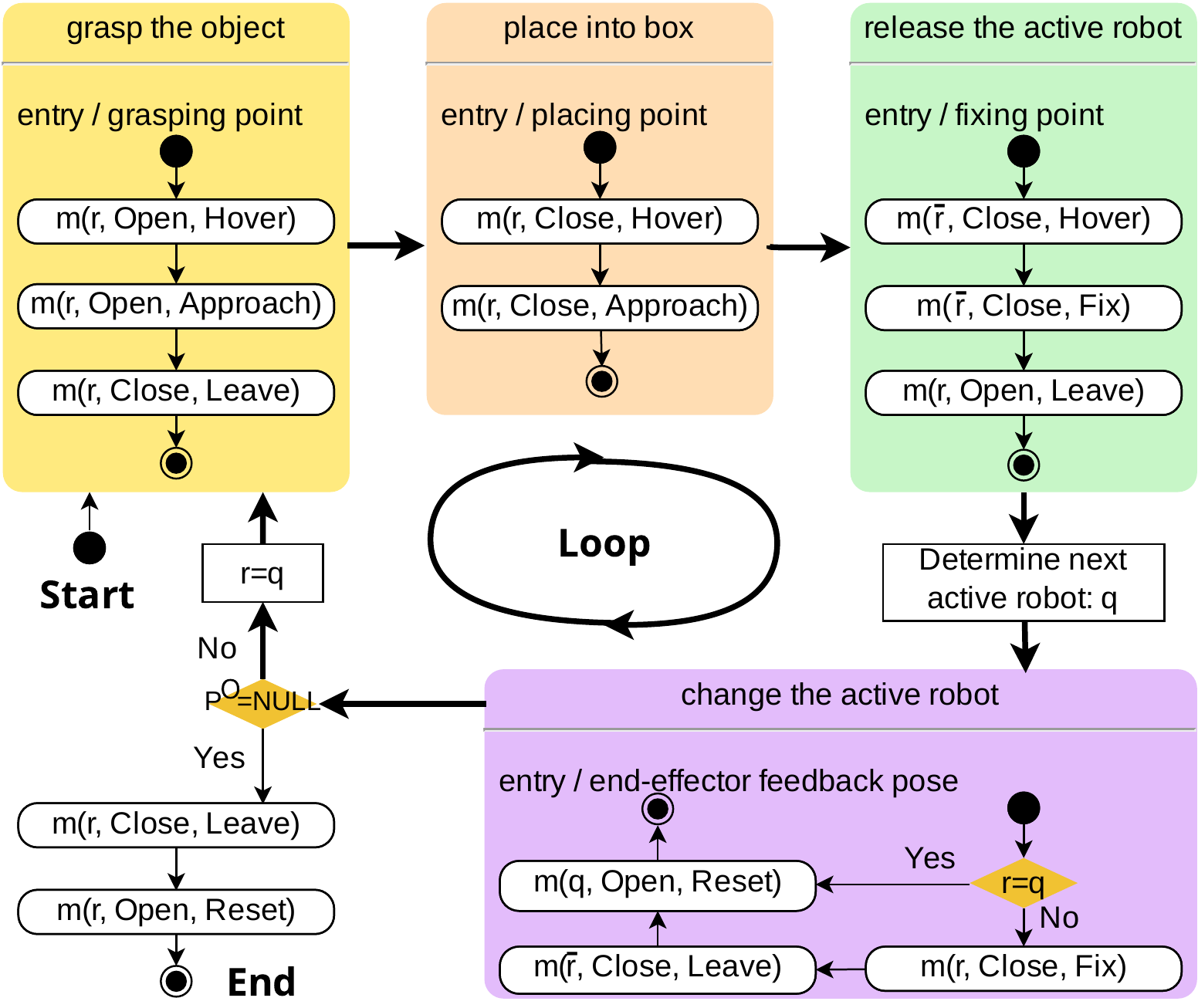}
		\caption{Action planner for packing task. High-level behaviors consist of robotic movements, e.g., grasping the object, placing it into the box, releasing the active robot after placing the object, and changing the identifier of the active robot. The results of reference point generator are inputs of behaviors as references. The specific action primitives determine the final target poses.}
		\label{state-machine}
		\vspace{-0.2cm}
	\end{figure}
	
	The proposed state machine to automatically pack the long linear elastic object has one periodic action planner loop (depicted in Fig. \ref{state-machine}), which is iterated while monitoring the object's state until the task is completed.
	This sequence of actions is performed by colaborative robotic arms, identified as \textit{Left} and \textit{Right} (see Fig. \ref{experimental setup}) that can alternate between an active role and an assitant role. 
	The former is in charge of grasping and placing the object into the box; The latter is in charge of immobilizing it while the arms change roles.
	
	Our method uses a collection of robot roles $R = \{ r, \overline r\}$, where $r$ specifies which robot takes up the active packing role in a given cycle of the action planner loop.
	The identifier $r=\textit{Left}/\textit{Right}$ is automatically determined based on the proximity of $\mathbf P^O$ to either the \textit{Left} or \textit{Right} robot.
	The assistant arm at the same cycle is denoted as $\overline{r}$, which for our dual-arm configuration, it simply represents the opposite arm, e.g., for $r=\textit{Right}$,  $\overline{r}=\textit{Left}$.

	The proposed state machine in Fig. \ref{state-machine} is composed of two layers.
	The first layer contains four high-level behaviors, namely, grasp the object, place it into the box, release the active robot, and change the active robot.
	The inputs of these high-level behaviors are the reference points $\mathbf p^G$, $\mathbf p^L$ and $\mathbf p^F$, and the reference poses $\mathbf x^*$. 
	The second layer contains several low-level robot movements; These are modelled as elements in the collection of states:
	\begin{equation}
	    S = \{ s : s = m(R, G, A) \} 
	\end{equation}
	where the triple $m(R, G, A)$ defines the robot movements as a sequence of the following two commands: (i) First, the active/assistant robot $r/\overline{r}\in R$ performs the gripper action primitive $g_i\in G$; (ii) Then, the robot performs the end-effector action primitive $a_j\in A$.
	
	The result of the robot movement $m(R, G, A)$ corresponding to each state $s$ is evaluated with the trasition function:
	\begin{equation}
		T(s)=
		\left\{
		\begin{array}{l}
			1,\,\, \textrm{once the robot completes the movement}, \\
			0,\,\, \textrm{otherwise.}\\
		\end{array}
		\right.
	\end{equation}
	
	The proposed action planner stops when no object points are detected outside the box. The packing task succeeds when the shape difference $e$ converges to $e^*$.
	
	\section{Results}\label{section: results}
	\subsection{Experimental Setup}\label{section: result: experiments}
	\begin{table}[t!]
		\centering
		\caption{properties of the objects in experiments}
		\begin{tabular}{ ccc }
			\toprule
			Material & Density (kg/m$^3$) & Young's Modulus (MPa)\\
			\midrule
			Natural Latex (NL) & 67.23 & 0.032\\
			Polyurethane Foam (PUF) & 38.76 & 0.185\\
			Silicone Foam (SCF) & 62.50 & 0.325\\
			Polyethylene Foam (PEF) & 16.17 & 0.992\\
			\bottomrule
		\end{tabular}
		\label{properties}
	\end{table}
	
	\begin{figure}[t!]
		\centering
		\includegraphics[width=\columnwidth]{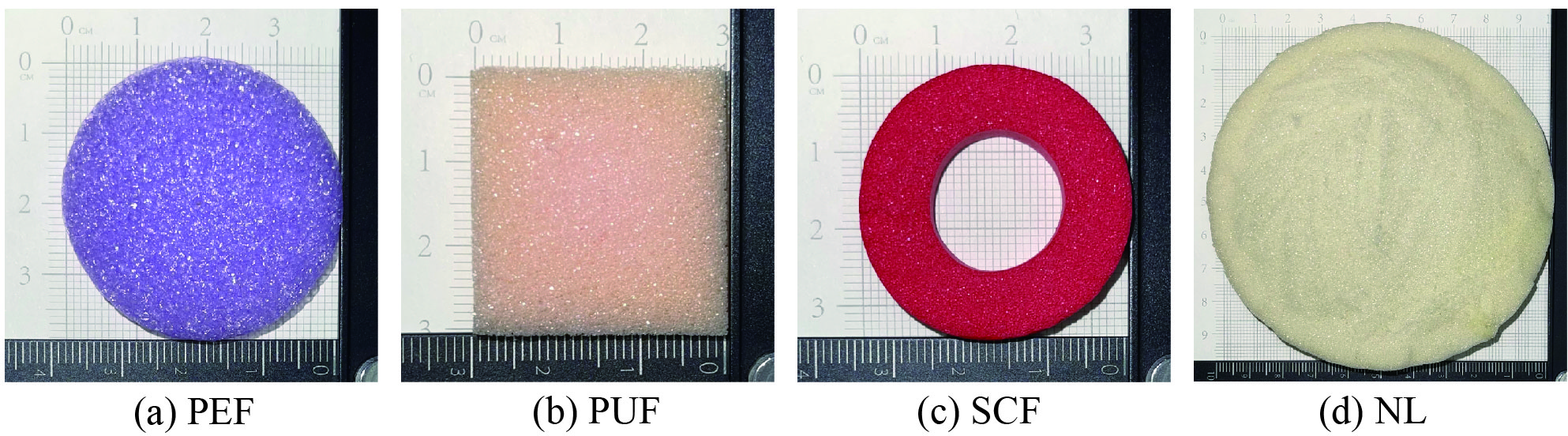}
		\caption{The widths/diameters $d_O$ and cross-sections of objects made of different materials. The measurements of $d_O$ are 38.0 mm, 30.0 mm, 34.0 mm, and 98.0 mm. The cross-sections are circle, square, ring, and circle.}
		\label{material-list}
	\end{figure}
	
	We conduct an experimental study to validate the proposed method.
	Fig. \ref{experimental setup} shows the developed experimental platform, which is composed of two 6-DOF robot manipulators (UR3) equipped with active grippers (Robotiq) that drive customized object grasping fixtures, and a top-view LiDAR camera (Intel RealSense L515) that captures real-time point clouds of the workspace.
	A table is placed between the two robot arms, with the packing box is rigidly attached to its surface. 
	The robotic arms are controlled with a Linux-based PC (running Ubuntu 16.04), with ROS and RViz used for communication and visualization \cite{ros}.
	Image processing is performed with the OpenCV libraries \cite{opencv_library}.

	\begin{figure}[t!]
		\centering
		\includegraphics[width=\columnwidth]{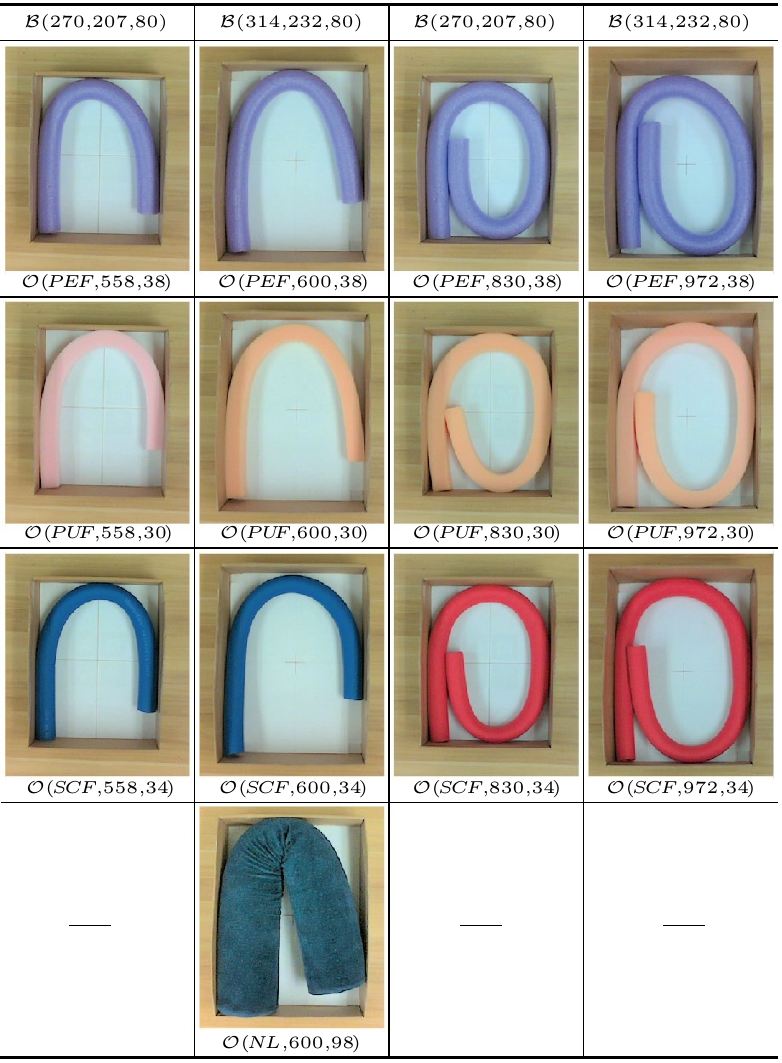}
		\caption{Packing 13 objects $\mathcal{O}(\eta, l_O, d_O)$ of different lengths into two boxes. PEF: Polyethylene Foam, PUF: Polyurethane Foam, SCF: Silicone Foam, NL: Natural Latex.}
		\label{case-list}
	\end{figure}

	To test the robustness of our method for packing LEOs, we use 13 objects with different elastic properties, cross-section shapes, and object lengths. 
	The density and Young's modulus of the object materials are listed in Table \ref{properties}; 
	The cross-sections, the widths and diameters of the objects are shown in Fig. \ref{material-list}; The 13 objects and its lengths are shown in Fig. \ref{case-list}.
	Twelve of these linear elastic objects are made of three materials: polyethylene foam (PEF), polyurethane foam (PUF), and silicone foam (SCF); These objects have four lengths: 558 mm, 600 mm, 830 mm, and 972 mm. 
	The thirteenth object is a pillow made of natural latex (NL) with a length of 600 mm.
	The objects in this study are all packed into boxes of two different sizes, viz. $\mathcal{B}(270,207,80)$ and $\mathcal{B}(314,232,80)$ (given in mm units).

	
	\begin{table}
		\centering
		\caption{Accuracy of geometric property estimation}
		\begin{tabular}{ ccc }
			\toprule
			Object & Length (\%) & Width/Diameter (\%)\\
			\midrule
			$\mathcal{O}(PEF, 558, 38)$ & 98.08 $\pm$ 1.68 & 93.16 $\pm$ 6.32\\
			$\mathcal{O}(PEF, 600, 38)$ & 97.73 $\pm$ 1.23 & 93.95 $\pm$ 4.74\\
			$\mathcal{O}(PEF, 830, 38)$ & 98.41 $\pm$ 1.00 & 91.05 $\pm$ 4.47\\
			$\mathcal{O}(PEF, 972, 38)$ & 98.80 $\pm$ 0.81 & 91.58 $\pm$ 5.26\\
			$\mathcal{O}(PUF, 558, 30)$ & 97.83 $\pm$ 1.67 & 91.34 $\pm$ 7.33\\
			$\mathcal{O}(PUF, 600, 30)$ & 97.77 $\pm$ 1.60 & 96.11 $\pm$ 6.07\\
			$\mathcal{O}(PUF, 830, 30)$ & 98.46 $\pm$ 1.22 & 98.08 $\pm$ 5.02\\
			$\mathcal{O}(PUF, 972, 30)$ & 98.91 $\pm$ 0.83 & 98.67 $\pm$ 6.10\\
			$\mathcal{O}(SCF, 558, 34)$ & 97.80 $\pm$ 1.16 & 93.82 $\pm$ 5.59\\
			$\mathcal{O}(SCF, 600, 34)$ & 98.18 $\pm$ 1.23 & 88.82 $\pm$ 7.35\\
			$\mathcal{O}(SCF, 830, 34)$ & 98.83 $\pm$ 0.99 & 87.65 $\pm$ 7.65\\
			$\mathcal{O}(SCF, 972, 34)$ & 98.89 $\pm$ 0.85 & 92.64 $\pm$ 6.47\\
			$\mathcal{O}(NL, 600, 98)$ & 99.27 $\pm$ 1.90 & 96.22 $\pm$ 3.16\\
			\bottomrule
		\end{tabular}
		\label{tab: initialization}
	\end{table}
	
	\subsection{Vision-Based Computation of the Objects' Geometry}
	We validate the accuracy of the model \eqref{equ:geometry} describing the objects' geometry (i.e., the length $l_O$ and width $d_O$) by collecting 10 measurements of their initial configuration over the table (similar to the one depicted in Fig. \ref{experimental setup}) and comparing the calculated dimensions with the ground truth, see Table \ref{tab: initialization}.
	The length $l_O$ is computed from the ordered point cloud $\mathbf{P}^O$, wheras $d_O$ is computed from the raw point cloud.
	The results in Table \ref{tab: initialization} show that the estimated object length $l_O$ and width $d_O$ are slightly smaller than the ground truth, which is caused by the discretization of the continuous objects' arc-length and the partial view of their surface.
	This, however, does not affect the proposed manipulation strategy, as demonstrated in the experimental results that follow.

	\subsection{Similarity Analysis of the Reference Template}
	\begin{figure}[t!]
		\centering
		\includegraphics[width=\columnwidth]{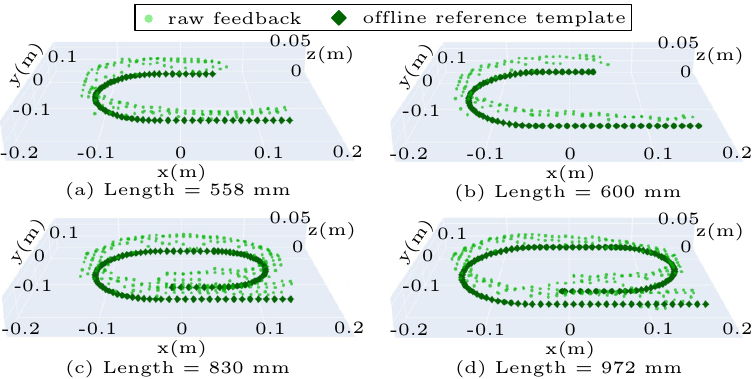}
		\caption{The comparison of raw feedback and the offline reference template.}
		\label{points}
	\end{figure}

	\begin{table}[t!]
		\centering
		\caption{performance of the method in packing tasks}
		
		\begin{tabular}{ cccc }
			\toprule
			Object & Mean $\mu$ & Variance $\sigma^2$ & 3-$\sigma$ Confidence Interval\\
			\midrule
			$\mathcal{O}(PEF, 558, 38)$ & 17.5 & 0.299 & 99.79\\
			$\mathcal{O}(PEF, 600, 38)$ & 18.0 & 0.411 & 99.84\\
			$\mathcal{O}(PEF, 830, 38)$ & 18.1 & 1.087 & 99.87\\
			$\mathcal{O}(PEF, 972, 38)$ & 18.5 & 0.874 & 99.42\\
			$\mathcal{O}(PUF, 558, 30)$ & 12.7 & 0.540 & 97.48\\
			$\mathcal{O}(PUF, 600, 30)$ & 15.1 & 0.078 & 98.61\\
			$\mathcal{O}(PUF, 830, 30)$ & 16.0 & 0.300 & 97.85\\
			$\mathcal{O}(PUF, 972, 30)$ & 15.1 & 0.360 & 98.03\\
			$\mathcal{O}(SCF, 558, 34)$ & 19.2 & 2.049 & 99.64\\
			$\mathcal{O}(SCF, 600, 34)$ & 19.6 & 0.130 & 99.73\\
			$\mathcal{O}(SCF, 830, 34)$ & 17.0 & 1.120 & 98.89\\
			$\mathcal{O}(SCF, 972, 34)$ & 20.3 & 1.435 & 99.85\\
			$\mathcal{O}(NL, 600, 98)$ & 50.9 & 1.984 & 99.94 \\
			\bottomrule
		\end{tabular}
		\label{tab: performance}
	\end{table}
	
	To verify if the designed $Spiral$ shape is able to match the desired object in boxes, we compute the similarity between the point clouds of the reference template and the raw feedback of the object. 
	To this end, we compute the set of minimum Euclidean distance between every point in $\mathbf{P}$ to the points in $\mathbf{P^*}$as follows:
	\begin{equation}
		D=\{ \min_j \| \mathbf{p}_i - \mathbf{p}_j^* \|_2: \mathbf{p}_i \in \mathbf{P}, \mathbf{p}_j^* \in \mathbf{P}^* \}
		\label{Euclidean}
	\end{equation}
	If the shape of the packed object matches $Spiral$ well, $D$ follows a Gaussian distribution $N(\mu,\sigma^2)$, with mean $\mu$ equal to the radius or half-width of the object $\mu\approx\frac{d_O}{2}$, and standard deviation $\sigma\approx 0$.
	The average value of the set $D$ is equal to the error $e_{in}$.
	Fig. \ref{points} presents the raw feedback point clouds $\mathbf P$ and the offline reference template $\mathbf P$.
	A statisticcal analytsis of the similarity is shown in Table \ref{tab: performance}, which demonstrates thatthe mean distances $\mu\approx\frac{d_O}{2}$ and the variances $\sigma^2$ are small. 

	\subsection{Generation of Reference Points}\label{section: result: target planning}
	\begin{figure}[t!]
		\centering
		\includegraphics[width=\columnwidth]{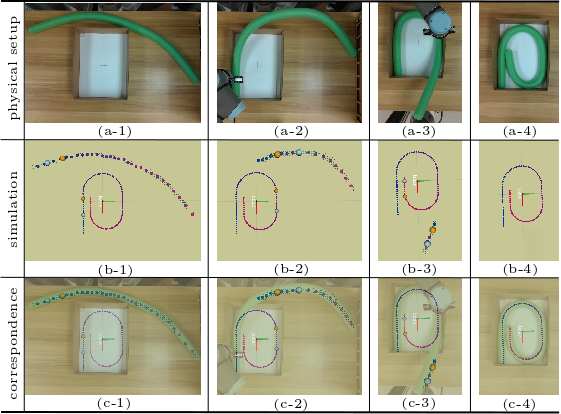}
		\caption{Reference point generator of action planner. The light green points are raw feedback points $\mathbf{P}$. Blue indicates the start, red indicates the end, and the gradients of colors indicate the order of points. The grasping points and placing points are orange, and the fixing points are light-blue. Blue indicates the start and red indicates the end, and the gradients of colors indicate the order of points.}
		\label{path-planning-result}
	\end{figure}
	In this section, we take $\mathcal{O}(PEF, 972, 38)$ as an example.
	The constant distance from $\mathbf{p}^{L}$ to $\mathbf{j}_B$ is set as $\delta_l = 50$ mm, and the distance from $\mathbf{p}^{F}$ to $\mathbf{p}^{L}$ as $\delta_f = 100$ mm.
	Based on these parameteres, the generator automatically computes the placing points $\mathbf p^L=\mathbf p_k^*$ and grasping points $\mathbf p^G=\hat{\mathbf p}_k$, for and index $k=\{26,64,110\}$; The fixing points $\mathbf p^F$ are determined within each cycle based on the location of the robot relative to each other. 
	Fig. \ref{path-planning-result} depicts the reference template and the ordered skeleton, where $\mathbf p^G$ and $\mathbf p^L$ are represented by organge points, whereas $\mathbf p^F$ by light-blue points. 
	In this figure, we can see how the active robot grasps the object at $\mathbf p^G$ (orange points on the object) and placed it at $\mathbf p^L$ (corresponding orange points at the template).
	The figure also shows how the assistant robot fixes the object by pushing it at $\mathbf p^F$ (corresponding light-blue point at the reference template), which enables the active robot to be released and conduct the next action.
	
	\subsection{Shape Difference}
	\begin{figure}[t!]
		\centering
		\includegraphics[width=\columnwidth]{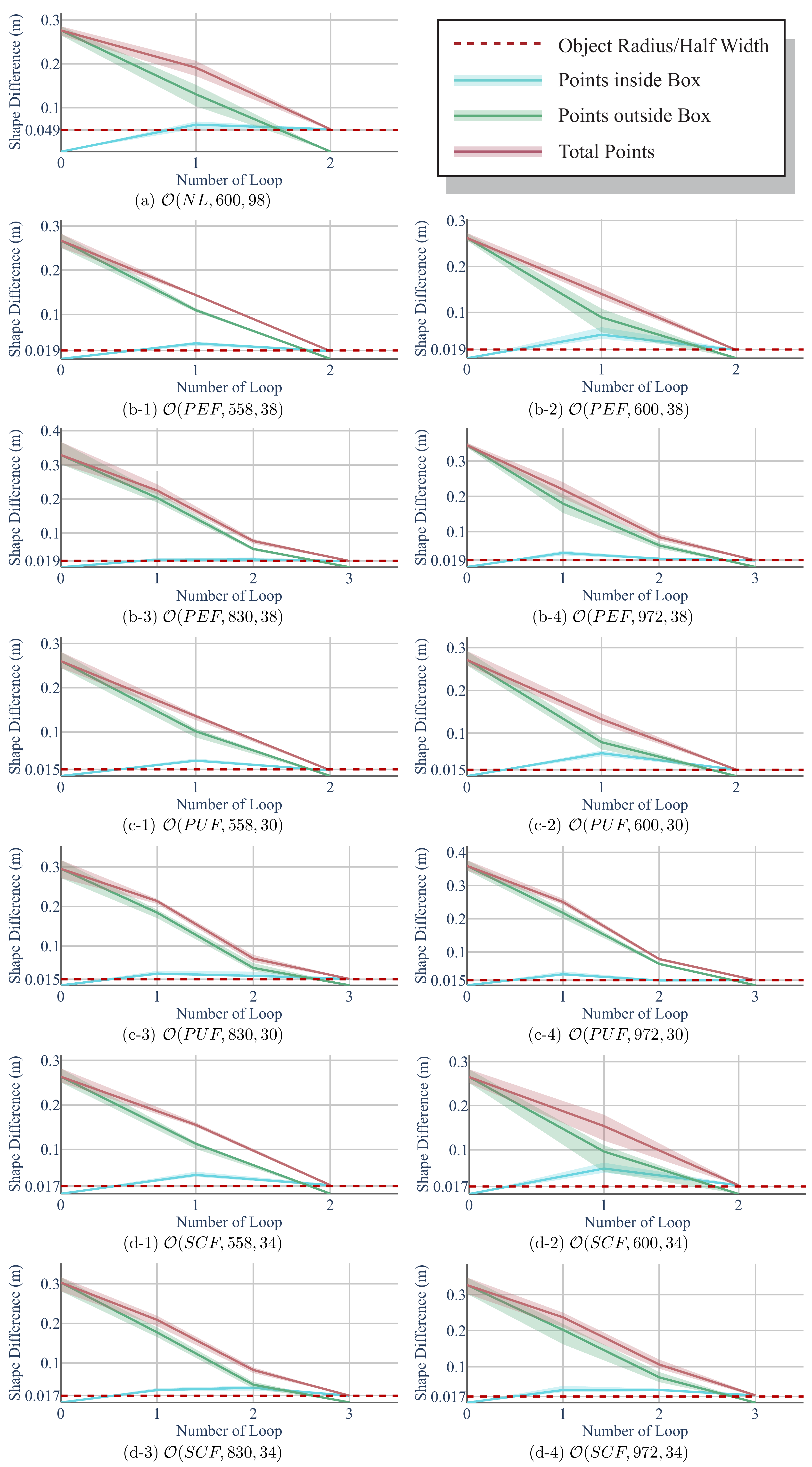}
		\caption{The shape differances (inside box, outside box, and total) during manipulation.}
		\label{shape-difference}
	\end{figure}
	
	This section validates the performance of the proposed automatic packing method with the 13 different objects shown in Fig. \ref{case-list}; Each experiment is conducted 10 times\footnote{\href{https://youtu.be/ZGJcRE2nqBc}{https://youtu.be/ZGJcRE2nqBc}}.
	To quantify the progress and accuracy of the packing task, we compute the shape difference with visual feedback; This metric is only computed at the beginning of every loop, as there are occlusions and noisy points when the robots are moving. 
	The blue, green, and red solid curves shown in Fig. \ref{shape-difference} respectively represent the errors $e_{in}$, $e_{out}$ and $e$ that are obtained from ten automatic packing experiments.
	The red dashed line represents the errors' ideal value $e^* = \frac{d_O}{2}$. 
	The blue curves start from zero when the object is completely outside the box before the automatic manipulation, and converge to $e^*$ after the object has been fully packed. 
	The green curves start from large initial values when the objects lie on the table with an underformed shape, and converge to zero when there are no points outside of the box after packing has been completed.
	The red curves, which represent the weighted average of the blue and green curves, start from the same initial values as the green curves, and monotonically decrease to $e^*$.
	These results quantitatively demonstrate that the proposed manipulation strategy can successfully deform and manipulate various types of LEOs into compact boxes.
	
	\subsection{High-Level Behaviors Constructed with Action Primitives}
	
	\begin{figure}[t!]
		\centering
		\includegraphics[width=\columnwidth]{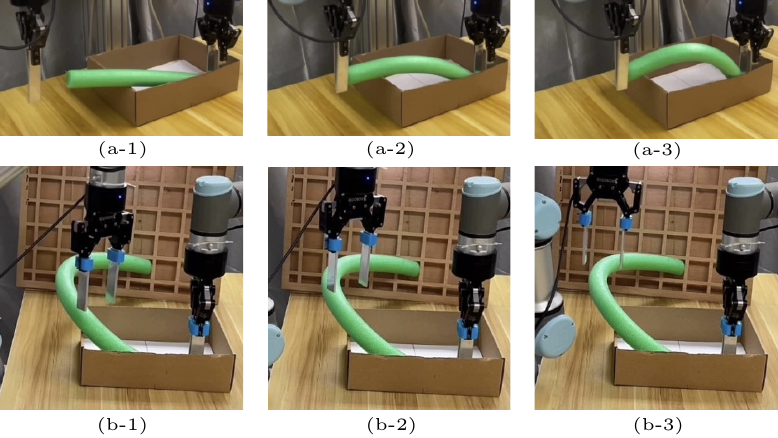}
		\caption{Robotic movements with two modes of \textit{Hover}. In (a-1)--(a-3), the robot moves at the constant height so that it touches the object causing the failure of grasping. In (b-1)--(b-3), the robot smoothly moves along the object with the following algorithm.}
		\label{collision}
	\end{figure}
	
	\begin{figure}[t!]
		\centering
		\includegraphics[width=\columnwidth]{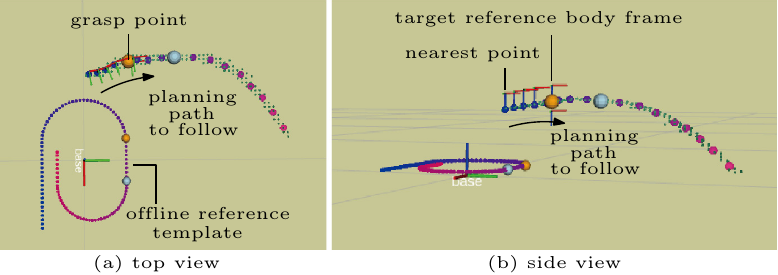}
		\caption{Path planning for robots following the object without collision.}
		\label{follow-path}
	\end{figure}
	
	Generally, it is sufficient for the robots to move at a constant height to avoid collisions with the edge of the cuboid-shaped bin. 
	However, as the ordered skeleton $\mathbf P^O$ is higher than the box, this constant height may produce collisions with the object, as shown in Fig. \ref{collision} (a-1)--(a-3).
	To deal with this problem, we proposed the \textit{Hover} action primitive, which guides the robot to move to the nearest point above the object, then, to move along the object's curvature until the gripper reaches the target grasping point $\mathbf p^G$, see Fig. \ref{follow-path}.
	This valuable of action primitive enables to successfully grasp LEOs with complex bent geometries, while avoiding collisions with them, as demonstrated in Fig. \ref{collision} (b-1)--(b-3).

\begin{figure}[t!]
		\centering
		\includegraphics[width=\columnwidth]{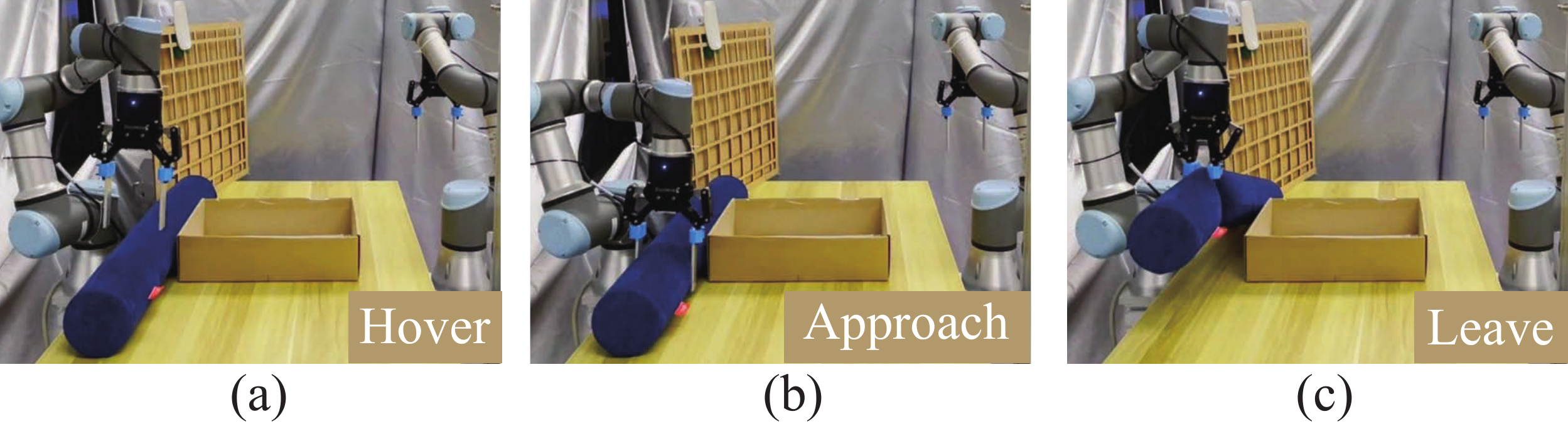}
		\caption{A grasping manipulation (Loop 1) is composed of action primitives given a grasping point $\mathbf{p}^{G}_1$: (a) $m (\textit{Left}, \textit{Open}, \textit{Hover})$, (b)$ m (\textit{Left}, \textit{Open}, \textit{Approach})$, (c) $m (\textit{Left}, \textit{Close}, \textit{Leave})$.}
		\label{ap-grasping}
	\end{figure}

\begin{figure}[t!]
		\centering
		\includegraphics[width=\columnwidth]{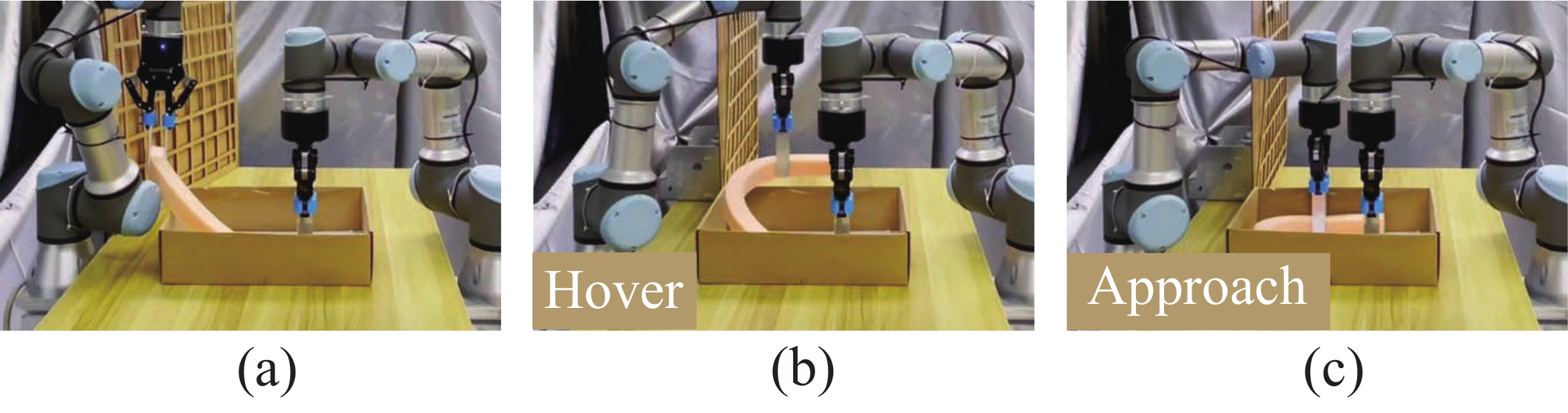}
		\caption{A placing manipulation (Loop 2) is composed of action primitives given a grasping point $\mathbf{p}^{L}_2$: (a) initial state, (b) $ m (\textit{Left}, \textit{Close}, \textit{Hover})$, (c) $m (\textit{Left}, \textit{Close}, \textit{Approach})$.}
		\label{ap-locating}
	\end{figure}

\begin{figure}[t!]
		\centering
		\includegraphics[width=\columnwidth]{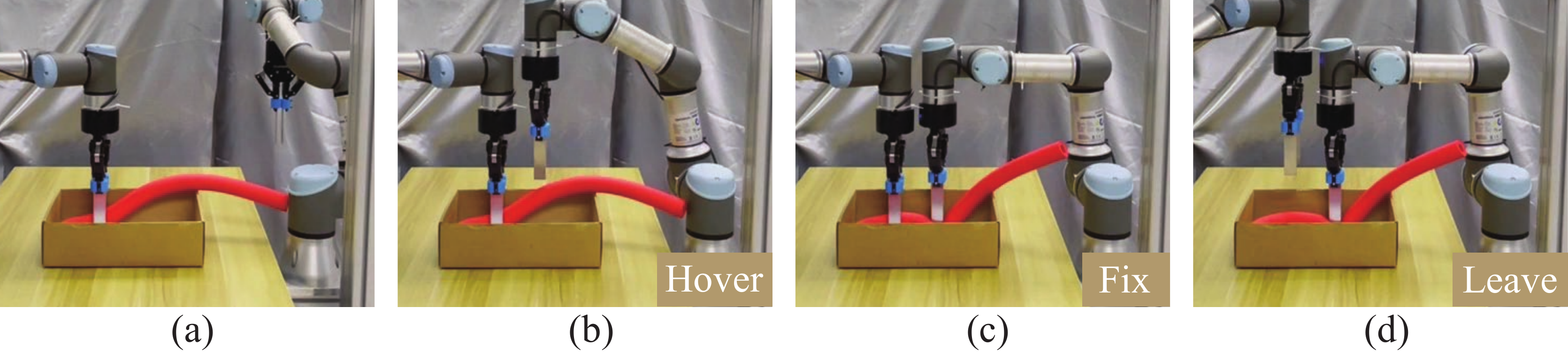}
		\caption{The assistant robot help to release the active robot that is grasping the object after placing it in the box (Loop 2) given a fixing point $\mathbf{p}^{F}_2$: (a) initial state, (b)$ m (\textit{Right}, \textit{Close}, \textit{Hover})$, (c) $m (\textit{Right}, \textit{Close}, \textit{Approach})$, (d) $m (\textit{Left}, \textit{Open}, \textit{Leave})$.}
		\label{ap-releasehand}
	\end{figure}

\begin{figure}[t!]
		\centering
		\includegraphics[width=\columnwidth]{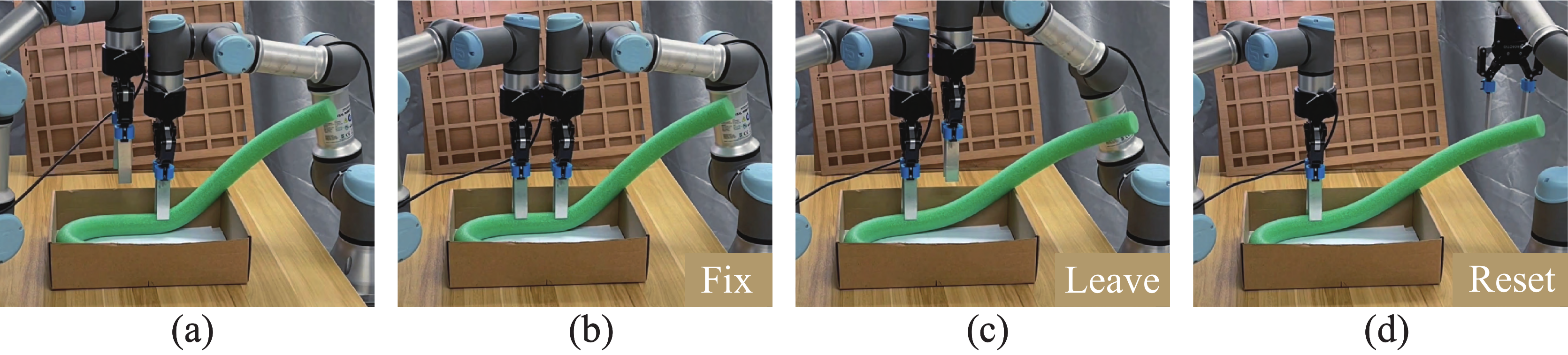}
		\caption{Changing hand manipulation (Loop 2) is composed of action primitives conducted by both robots concerning their current positions: (a) initial state, (b) $m (\textit{Left}, \textit{Close}, \textit{Fix})$, (c) $m (\textit{Right}, \textit{Close}, \textit{Leave})$, (d) $m (\textit{Right}, \textit{Open}, \textit{Reset})$.}
		\label{ap-changehand}
	\end{figure}

	The four high-level behaviors (i.e., grasp the object, place it into the box, release the active robot, and change the active robot) are shown in Fig. \ref{ap-grasping}--\ref{ap-changehand}. 
	The experiment in Fig. \ref{ap-grasping} shows the how the robot autonomously grasps the object and reaches a safe height from the table.  
	This figure shows that the \textit{Left} robot performs \textit{Hover} above $\mathbf{p}^{G}$, then \textit{Approach} towards the grasping point $\mathbf{p}^{G}$ with an open gripper, and finally \textit{Close} and \textit{Leave} from $\mathbf{p}^{G}$ towards the initial height $z(t_0)$.
	The second high-level behavior is depicted in 	Fig. \ref{ap-locating}. 
	The purpose of these sequence of actions is to deform and place the grasped object at a specific position within the box.
	The figure shows the initial configuration where the object is grasped by the \textit{Left} robot and the inside-the-box part is fixed by the \textit{Right} robot; Then, the \textit{Left} robot performs \textit{Hover} and \textit{Approach} towards $\mathbf{p}^{L}$, while holding the object.
	
	The third high-level behavior (release the active robot) is depicted in Fig. \ref{ap-releasehand}. 
	The purpose of these movements is to fix the object's shape while the active robot that is holding the object opens its gripper.
	The figure shows the initial configuration where the object (already inside the box) is grasped by the \textit{Left} robot; Then, the \textit{Right} robot performs \textit{Hover} and \textit{Approach} $\mathbf{p}^{F}$ with a closed gripper. 
	The \textit{Left} robot \textit{Open} its gripper, \textit{Leave} the object, and returns to the initial height $z(t_0)$.
	The fourth high-level behavior (change the active robot) is depicted in Fig. \ref{ap-changehand}.
	The purpose of these movements is to switch the active robot's identifier from \textit{Left} to \textit{Right}, this, in preparation for the \textit{Right} robot to conduct the next grasp task. 
	The figure shows the initial state where the \textit{Left} robot is free, and the right robot is performing \textit{Fix} onto the object; Then, the \textit{Left} robot performs the \textit{Fix} action while the \textit{Right} robot performs \textit{Leave} and then \textit{Reset} to return to its initial position, which completes one cycle of the action planner loop.

\begin{figure*}[t!]
		\centering
		\includegraphics[width=18cm]{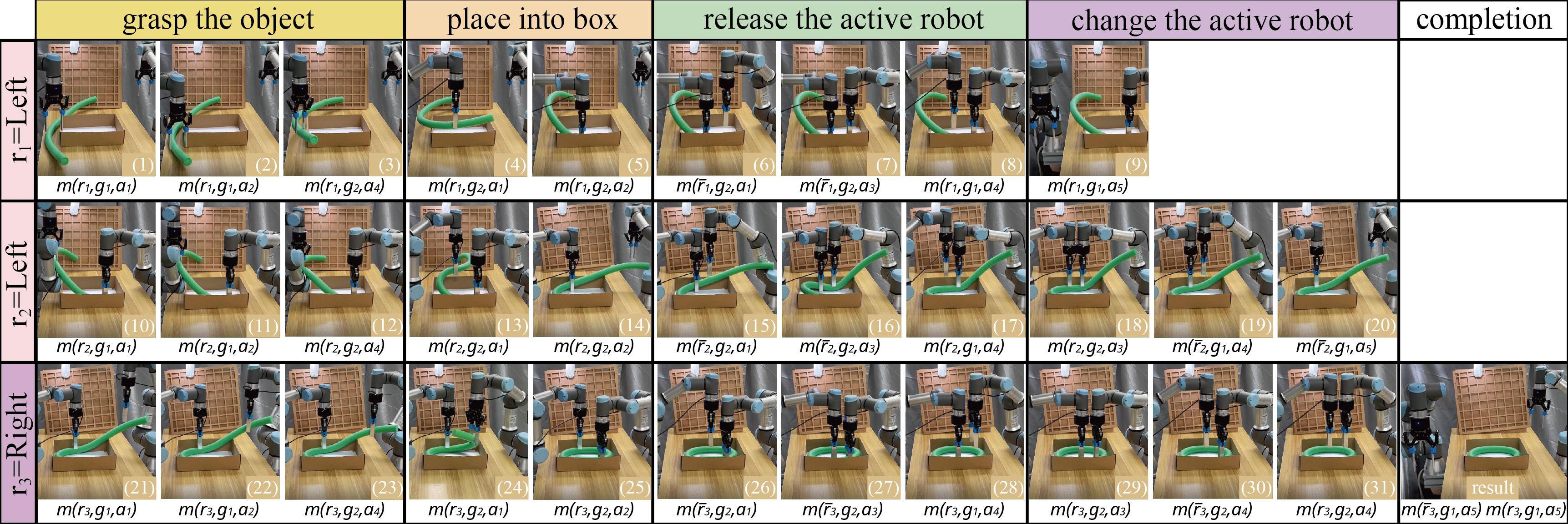}
		\caption{Experiment process and action primitives of packing $\mathcal{O}(PEF, 972, 38)$. The rows represent three action planner loops. The columns represent the behaviors of robots. The thumbnails demonstrate robotic movements. The first two loops are mainly executed by the left arm (grasping and placing) and assisted by the right arm (fixing). The third loop is mainly executed by the right arm (grasping and placing) and assisted by the left arm (fixing). }
		\label{action primitives}
	\end{figure*}

	We take $\mathcal{O}(PEF, 972, 38)$ as a representative example to demonstrate the performance of the method. 
	Fig. \ref{action primitives} depicts the complete process of the packing task, which consists of three cycles, corresponding to the three rows; Each thumbnail in the figure presents a movement $m(R,G,A)$ conducted by the robot arms.
	The periodic nature of the action planner is illustrated by the fact that the three cycles share the same first three high-level behaviors, viz. grasp the obejct, place into the box, and release the active robot. 
	The first and second cycles only differ in the fourth high-level behavior, i.e., change the active robot, as the active robots in these two cycles are the same (thus, there is no need to change the active robot).
	The packing process ends in the third cycle, where the object has been completely packed into the box.

	
	\section{Conclusions}\label{section: conclusions}
	
	In this work, we propose a complete method to pack long linear elastic objects into compact boxes. First, we design a hybrid geometric model including an online 3D-vision method and an offline reference template to tackle occlusions during packing manipulations under a single-view camera.
	Online 3D-vision method extracts objects' geometric information in real time.
	Offline reference template is generated based on a designed shape $Spiral$.
	The effectiveness of $Spiral$ is proved by the high similarity between the offline reference template and the shape of the packed object. 
	Then, we propose a method to preliminarily plan reference points for grasping, placing, and fixing. 
	Next, we propose an action planner to compose defined action primitives as high-level behaviors and achieve packing tasks by repeating a periodic action planner loop. 
	Finally, extensive experiments are conducted to verify the generality of our proposed method for various objects with different elastic materials, lengths, densities, and cross-sections. 
	
	Although the method is designed for packing tasks, the defined action primitives and the reference point generator method can be used in other manipulation tasks (e.g. object sorting, multiple objects assemblies, etc). 
	Also, the proposed perception method is able to work without markers and decrease computation time by extracting minimum geometic information of objects.
	A limitation of our method is that our perception method does not consider the situations where the object is outside the camera's view range. 
	A possible solution is to employ multi-view visual system to perceive the object. 

	For future work, we plan to explore the multi-view vision and to extend the framework to other comprehensive tasks involving more types of objects (e.g., rigid, elastic, articulated), as well as to optimize the packing to save space.
	Our team is currently working along this challenging direction.
	
	\ifCLASSOPTIONcaptionsoff
	\newpage
	\fi

	\bibliographystyle{IEEEtran}
	\bibliography{IEEEabrv,works_cited}

	\begin{IEEEbiography}
		[{\includegraphics[width=1in,height=1.25in,clip,keepaspectratio]{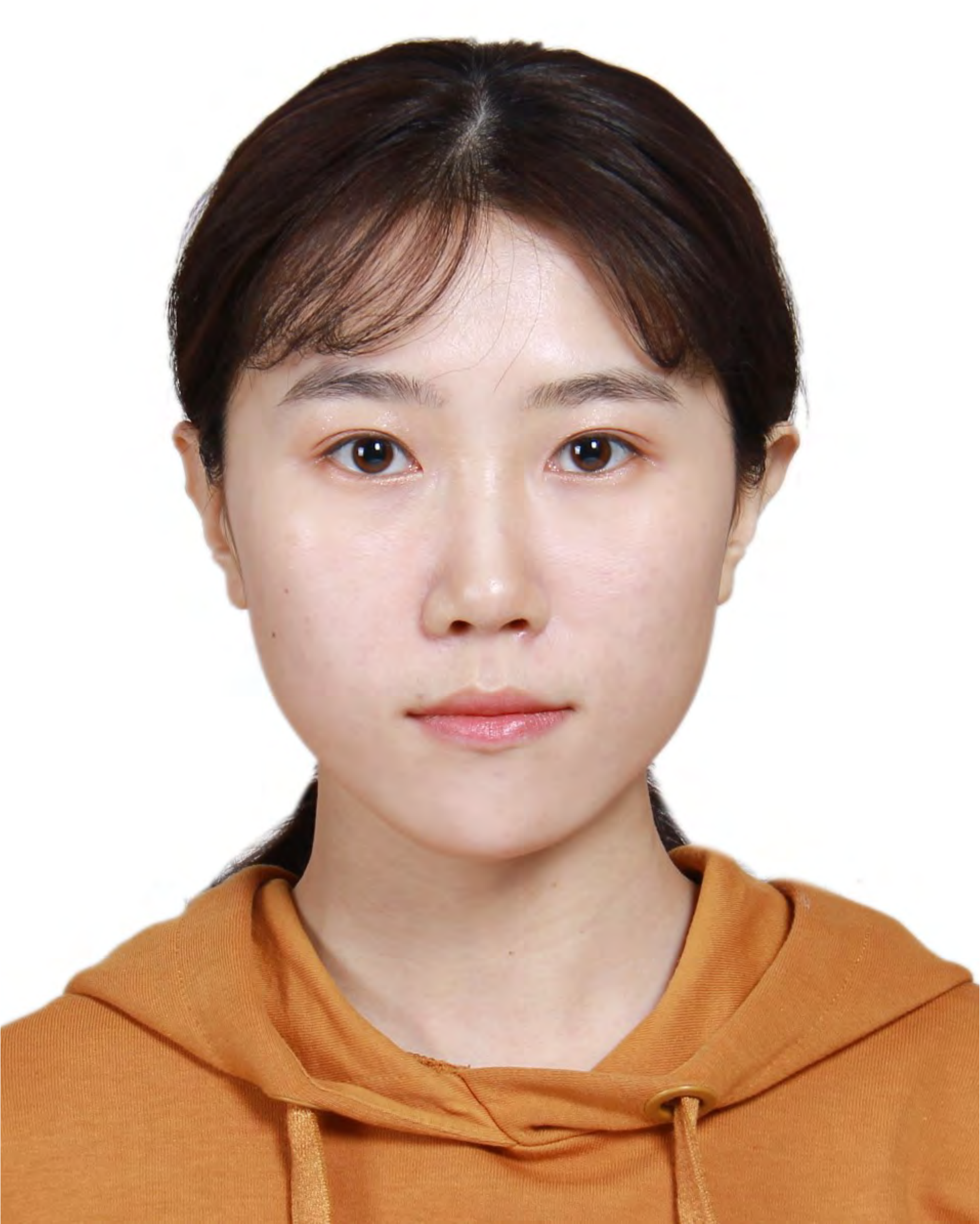}}] {Wanyu Ma} 
		received the B.S. and MA.Eng degrees in control science and engineering from Harbin Institute of Technology, Harbin, China, in 2016 and 2018, respectively. 
		She is currently working toward the Ph.D. degree in robotics with the Department of Mechanical Engineering, The Hong Kong Polytechnic University, Hong Kong.
		Her research interests include robotics, visual servoing, and deformable objects manipulation.
	\end{IEEEbiography}
	
	\begin{IEEEbiography}
		[{\includegraphics[width=1in,height=1.25in,clip,keepaspectratio]{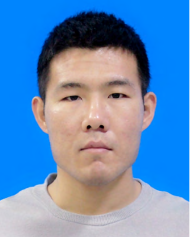}}] {Bin Zhang}
		received the B.S. degree from Beihang University, Beijing, China, in 2017, and the M.S. degree from China Academy of Space Technology, Beijing, China, in 2020.
		He is a Ph.D. student with the Department of Mechanical Engineering at The Hong Kong Polytechnic University. His research interests include multi-agent systems and control theory.
	\end{IEEEbiography}
	
	\begin{IEEEbiography}
		[{\includegraphics[width=1in,height=1.25in,clip,keepaspectratio]{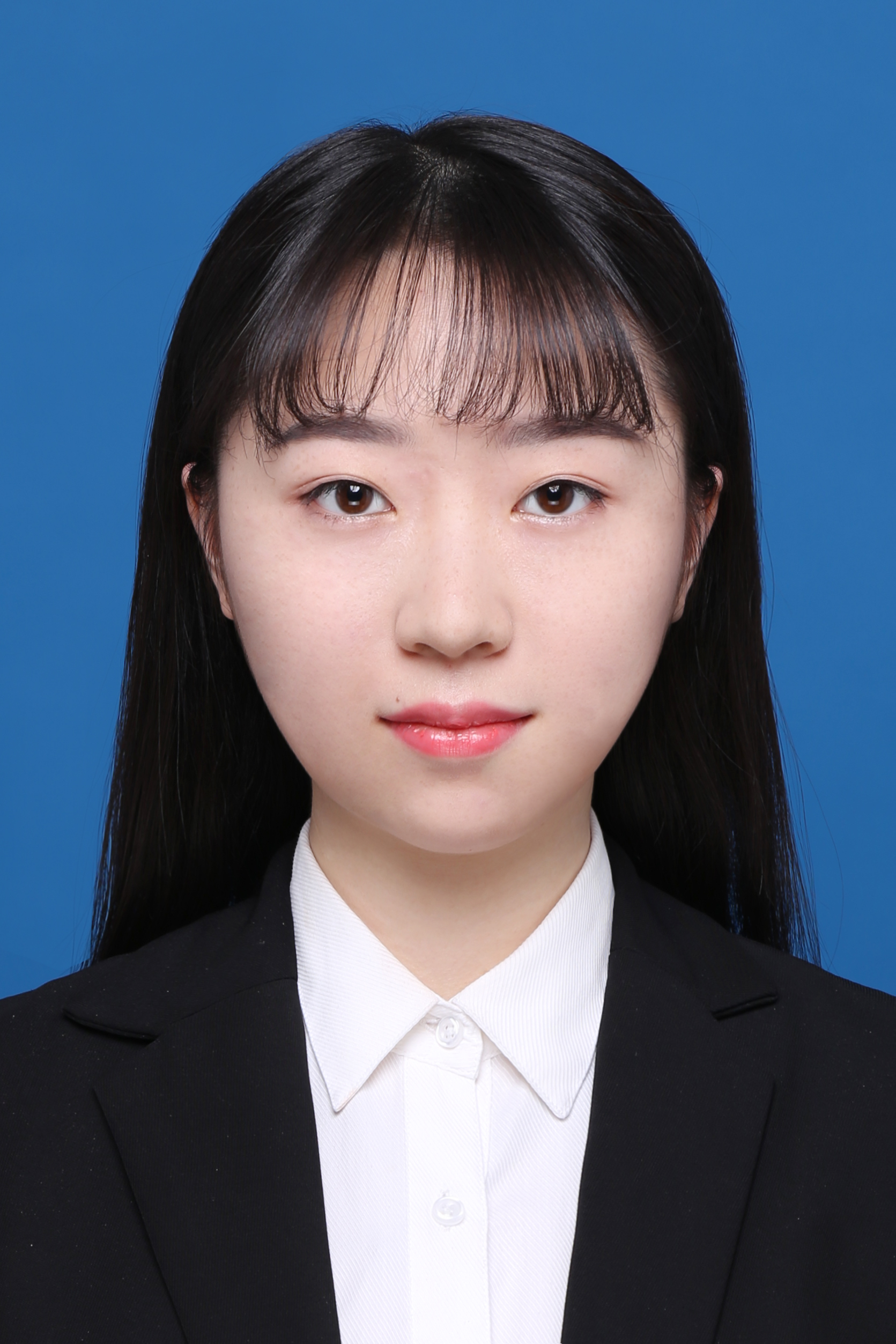}}] {Lijun Han}
		received the B.S. degree in automation from Tianjin University, Tianjin, China, in 2019. She is currently working toward the Ph.D. degree in control technology and
		control engineering with the Department of
		Automation, Shanghai Jiao Tong University,
		Shanghai, China.
		Her research interests include visual
		servoing, robot control, and surgical robots.
	\end{IEEEbiography}
	
	\begin{IEEEbiography}
		[{\includegraphics[width=1in,height=1.25in,clip,keepaspectratio]{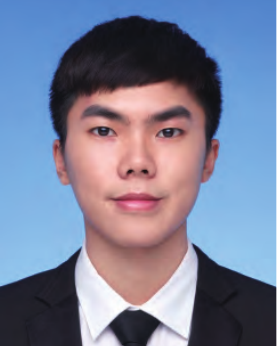}}] {Shengzeng Huo}
		received the B.S. degree in vehicle engineering from the South China University of Technology, Guangzhou, China, in 2019. 
		He is currently pursuing the Ph.D. degree in the Department of Mechanical Engineering from The Hong Kong Polytechnic University, Hong Kong. His research interests include bimanual manipulation, deformable object manipulation, and robot learning.
	\end{IEEEbiography}
	
	\begin{IEEEbiography}
		[{\includegraphics[width=1in,height=1.25in,clip,keepaspectratio]{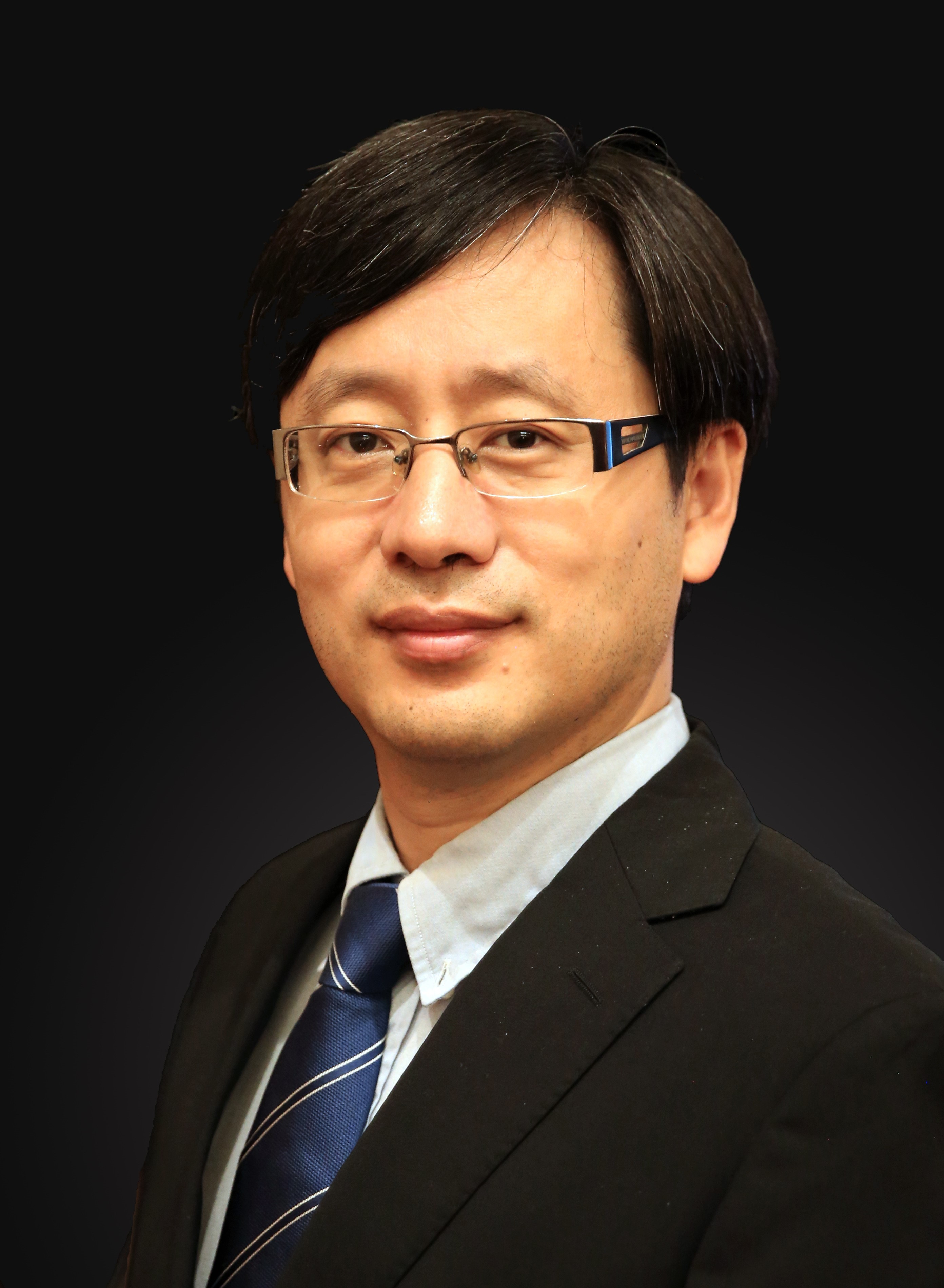}}] {Hesheng Wang}
		(Senior Member, IEEE) received the B.Eng. degree in electrical engineering from the Harbin Institute of Technology, Harbin, China, in 2002, and the M.Phil. and Ph.D. degrees in automation and computeraided engineering from The Chinese University of Hong Kong, Hong Kong, in 2004 and 2007, respectively.
		He is currently a Professor with the Department of Automation, Shanghai Jiao Tong University, Shanghai, China. His current research interests include visual servoing, service robot, computer vision, and autonomous driving.
		Dr. Wang is an Associate Editor for IEEE Robotics and Automation Letters, Assembly Automation and the International Journal of Humanoid Robotics, and Technical Editor for the IEEE/ASME TRANSACTIONS ON MECHATRONICS. From 2015 to 2019, he was an Associate Editor for the IEEE TRANSACTIONS ON ROBOTICS. He was the General Chair of the IEEE RCAR 2016 and the Program Chair of the IEEE ROBIO 2014 and IEEE/ASME AIM 2019.
	\end{IEEEbiography}
	
	\begin{IEEEbiography}
		[{\includegraphics[width=1in,height=1.25in,clip,keepaspectratio]{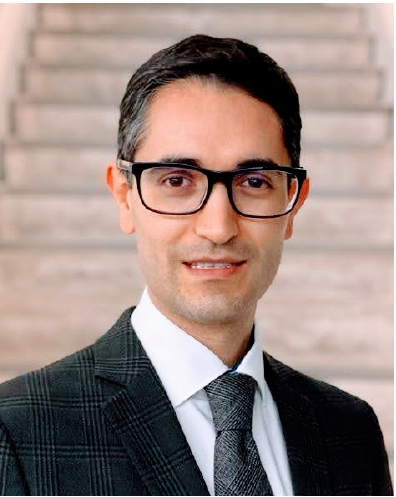}}] 	{David Navarro-Alarcon} (GS'06--M'14--SM'19) received the Ph.D. degree in mechanical and automation engineering from The Chinese University of Hong Kong, Shatin, Hong Kong, in 2014. 
		
		From 2014 to 2017, he was a Postdoctoral Fellow and then a Research Assistant Professor at the CUHK T Stone Robotics Institute, Hong Kong. Since 2017, he has been with The Hong Kong Polytechnic University, Kowloon, Hong Kong, where he is currently an Assistant Professor at the Department of Mechanical Engineering, and the Principal Investigator of the Robotics and Machine Intelligence Laboratory.
	His current research interests include perceptual robotics and control theory.
	\end{IEEEbiography}
	
\end{document}